\documentclass[11pt]{article}
\usepackage[margin=1in]{geometry}
\usepackage{natbib}
\usepackage{amsmath,amssymb,amsthm}
\usepackage{graphicx}
\usepackage{booktabs}
\usepackage{hyperref}
\usepackage{xcolor}
\usepackage{algorithm}
\usepackage{pifont}
\usepackage[ruled, linesnumbered, noend, algo2e]{algorithm2e}
\usepackage{multirow}
\usepackage{rotating}
\usepackage{siunitx}
\usepackage{caption}
\usepackage{subcaption}
\newcommand{\cmark}{\ding{51}}
\newcommand{\xmark}{\ding{55}}

\newcommand{\boldy}{\mathbf{y}}
\newcommand{\boldx}{\mathbf{x}}
\newcommand{\boldz}{\mathbf{z}}

\newcommand{\bAttn}{b_{\mathrm{attn}}}

\newcommand{\dmodel}{D_{\mathrm{emb}}}

\title{Observation Assimilation Transformer\\
Probabilistic Weather Analysis from Sparse Station Observations}

\author{Christian Donner$^{\star,1}$, Michele Cattaneo$^{\star,1}$, Maxim Samarin$^{1}$, \\ 
Francesco Zanetta$^{2}$, Ic\'iar Llor\'ens Jover$^{2}$, Gabriela Aznar Siguan$^{2}$, Michele Volpi$^{1}$ \\
\\
\small $^1$ Swiss Data Science Center, Eidgen\"ossische Technische Hochschule Z\"urich (ETHZ)
\\
\small $^2$ Federal Office of Meteorology and Climatology MeteoSwiss, Switzerland
\\
\small $^\star$ contributed equally
}
\begin{document}

\maketitle

\begin{abstract}
We present the Observation Assimilation Transformer (OAT), a geometry-aware neural process that corrects and refines a prior numerical weather prediction (NWP) field into a high-resolution posterior field by assimilating sparse observations. OAT is an encoder-decoder transformer that considers a set of observed locations and their NWP prior to condition predictions at a set of arbitrary unobserved query locations. Its sparse attention mechanism makes it scalable to thousands of station observations and depends on relative positions between locations, never on the absolute coordinates. The model allows for partially missing observations, predictions at arbitrary continuous locations without grid projection, and outputs are parametric distributions. A station-randomized training algorithm is developed with continuous ranked probability score (CRPS) loss. We test OAT on two datasets -- one at regional scale with hundreds of stations, and one at continental scale with thousands of stations. On these datasets OAT outperforms several baselines and shows well calibrated predictions. In an illustrative case study, OAT's downscaling capabilities are demonstrated by downscaling ERA5 ($\sim31$km) to COSMO ($1$km), where OAT achieves lower mean absolute error (MAE) and CRPS for temperature and precipitation on test stations compared to COSMO.
\end{abstract}
\section{Introduction}

Numerical weather prediction (NWP) models solve the governing equations of the atmospheric motions on high-performance computing (HPC) systems \citep{bauer2015quiet}, but their utility is fundamentally limited by the spatial resolution at which these equations can be practically solved. Their spatial and temporal discretization introduces errors and systematic biases through imperfect modelling of processes occurring at scales smaller than the model grid. 
Additionally, in the presence of complex topography, the spatial resolution of the NWP grid may be insufficient to accurately represent the underlying terrain over which the equations are solved. While increasing the horizontal spatial resolution can reduce some of these errors, doing so comes at a substantial increase of computational cost.
Fully data-driven alternatives to NWP models are currently achieving remarkable results at a fraction of the inference cost, often matching or surpassing NWP models on specific metrics. However their trustworthiness and ability to model extreme events is still, at the time of writing, an open question \citep{bonavita2024some}. 
A natural approach to address these limitations, is to reconcile NWP output with observations, which, despite being affected by measurement errors, are the best available estimate of the true atmospheric state.

Statistical postprocessing \citep{rasp2018neural} addresses the issue by learning a distribution that corrects systematic errors of raw NWP ensemble forecasts at station locations, conditioned on an NWP ensemble, geographic location and a set of predictors features derived from topography. Observations serve as training target, and if the learned model does not rely on station-specific embeddings, it can be transferred to arbitrary locations, under the assumption that topographic features alone are sufficient to describe the local processes. However, statistical postprocessing treats prediction locations independently, where neighboring stations are not used to inform predictions at nearby or related locations. Furthermore, observations are only used as training targets, meaning the model can be applied before any observation is available but fails to assimilate the actual realization of the target process at inference time.
Statistical downscaling \citep{rampal2024enhancing, mardani2025residual} addresses a complementary limitation; rather than correcting errors at station locations, it learns a mapping from a coarse resolution field, to fine-scale predictions by targeting high-resolution gridded references. This may or may not include coupled past observations \citep{vaughan2022convolutional,bruinsma2023autoregressiveconditionalneuralprocesses,sousa2026earthobservationembeddingseffective}. 
However, like statistical postprocessing, downscaling typically does not assimilate observations at inference time. In contrast, data assimilation \citep{carrassi2018data} explicitly combines model forecasts with newly available observations to refine the estimated atmospheric state at any given time, and is an integral part of the NWP pipeline. These approaches are complementary; statistical postprocessing focuses on correcting biases, statistical downscaling primarily enhances spatial resolution and data assimilation incorporates observations at inference time. 


We aim to combine elements from each paradigm into a data-driven framework that can correct NWP biases, generate higher resolution fields, and condition probabilistic predictions on observations. To this end, we present the \emph{Observation Assimilation Transformer} (OAT), a geometry-aware neural process combining:
\begin{itemize}
	\item{\bf NWP \& Topography} informed predictions, embedding both sources into a (learned) latent vector.
	\item{\bf Observation assimilated predictions} via a dedicated transformer architecture.
	\item{\bf No absolute positions; } attention only depends on relative geometry similar to~\cite{ashman2024translation}.
	\item{\bf Scalability to thousands of stations} via kNN-sparsified attention.
	\item{\bf Embedding partial observations} via a neural gating mechanism.
	\item{\bf A randomized training algorithm} using station observations as target.
\end{itemize}

After discussing related work in Section~\ref{sec:related}, we introduce the architecture and training procedure of OAT in Section~\ref{sec:methodology}. Then we introduce two datasets, one focusing on Switzerland and one whole Europe in Section~\ref{sec:experiments}. We show OAT's capabilities on these datasets and we compare it against several data-driven baselines, analyze the performance of OAT under different factors, show a downscaling example and the calibration of OAT to provide predictive distribution. Section~\ref{sec:discussion} concludes discussing limitations and future work. All code is made available at the \href{https://github.com/SwissDataScienceCenter/observation_assimilation_transformer}{Swiss Data Science Center Github}.

\section{Related Work}
\label{sec:related}

\subsection{  
Gaussian Processes and Data Assimilation Approaches}

For decades, researchers attempt to incorporate sparse observations into NWP fields to correct and downscale them. 
Traditionally, statistical methods such as linear correction models~\citep{burger1996expanded} and kriging methods~\citep{biau1999estimation} are used to estimate a correction function of the field by regression. Today machine learning methods are on the forefront of data-driven downscaling and data-assimilation.  \citet{zanetta2026efficient} use Gaussian Processes (GPs) for downscaling NWP wind fields, assimilating station observations. The GP prior's mean is parametrized by a neural network,  parsing NWP products and topography features.
The authors then propose to use deep kernel learning~\citep{wilson2016deep} to condition the GP posterior on station observation. 

\subsection{ML models for combining NWP products with spatial observations}
There have been various machine learning methods proposed for combining observational station data with gridded products. \citet{miralles2025observation} use graph neural networks (GNN) \citep{defferrard2016convolutional, kipf2016semi} for improving nowcasting. \citet{yang2024offgrid} use transformer and GNN architectures for forecasting at station level. While our work focuses more on spatial aggregation, it combines the transformer and GNN perspective of both works. We will rely on the transformer architecture, but can also view the proposed aggregation as belonging to the message-passing framework \citep{gilmer2017neural} in GNNs. MetNet-3 \citep{andrychowicz2023deep} combines gridded numerical, gridded observational, and sparse observational station inputs in a U-Net architecture trained to predict three types of targets, among which the sparse station input data densified onto a 4 km $\times$ 4 km grid. 
Statistical postprocessing methods such as the classic EMOS \citep{gneiting2005calibrated} or more recent deep learning approaches \citep{rasp2018neural, zanetta2023physics} correct NWP predictions at station locations using relationships learned from historical observational data, but do not condition on real-time observations at inference time.

\subsection{Neural processes}
An interesting model class that tries to address the scalability limitations of GPs are Neural Processes~\citep{garnelo2018neural} (NPs), but they come with arguably less consistent conditioning properties. For efficient processing of spatial data, convolutional neural processes~\citep{gordon2019convolutional} have been proposed and have also been applied to weather forecasting~\citep{vaughan2022convolutional}. In this work they also combine gridded NWP products with station data to predict temperature and precipitation at unseen locations. However, despite the elegant model construction that enables prediction anywhere, the main computation effectively happens at the resolution of the gridded data.

To assimilate sparse observations without gridded data, \citet{kim2019attentive}  developed the attentive neural processes. This model uses cross-attention to aggregate observations into predictions at the locations of interest. A problem with attention is that computation and memory scale quadratically with the number of observations. Hence, \citet{bao2024spatial} extended the model to spatial multi-attention conditional neural processes (SMACNP) to specifically update spatial information and uncertainty via different cross-attention channels. \citet{dumitrescu2026deep} have adopted the model for assimilation of daily weather variables, aggregating station observations and topography for Romania.

However, only relying on cross-attention, i.e. each query location aggregates over all observations, neglects certain information that the set of observations can provide. For example, if two observations contradict each other, the model should weigh them accordingly. This can be handled by the transformer neural process (TNP)~\citep{nguyen2022transformer}, exploiting self-attention between the observed points to update the information flow, before incorporating it into the cross-attention step. Yet, in the standard formulation, the attention uses the absolute positions to learn how locations are related. This is undesirable in weather prediction, because it could lead to memorizing specific frequent location-dependent patterns, without learning broader patterns that generalize across locations. To address this, researchers developed translation equivariant TNP (TE-TNP)~\citep{ashman2024translation} and its extensions~\citep{ashman2024approximately}. Furthermore, \citet{ashman2025gridded} addresses the scalability issue with a gridded TNP (although not translation equivariant) by encoding all observations (gridded and unstructured) on a regular grid, which then allows using Swin Transformers~\citep{liu2021swin} to learn efficiently global relations.

In conclusion, while significant effort has been made to develop ML models for data aggregation, there is no method that covers the full spectrum: a model that (i) aggregates information from NWP products, topography, and sparse observations, (ii) is scalable to thousands of station observations, (iii) and does not rely on absolute position (see Table~\ref{tab:comparison}). Our work closes this gap.

\begin{table}[h]
\centering
\small
\begin{tabular}{lccccc}
\toprule
 & NWP prior & Geo-attn & Station SA & Pred. anywhere & Scalable \\
\midrule
(Deep Kernel) GP~\citep{zanetta2026efficient} & \cmark & $\sim$ & \cmark    & \cmark & $\sim$ \\
ConvCNP~\citep{gordon2019convolutional}     & \cmark & \xmark & \xmark & \cmark & \cmark \\
TE-TNP~\citep{ashman2024translation, ashman2024approximately} & \xmark & \cmark & \cmark & \cmark & \xmark \\
Gridded TNP~\citep{ashman2025gridded}       & \xmark & \xmark & $\sim$ & \cmark & \cmark \\
SMACNP~\citep{bao2024spatial}  & \xmark & $\sim$ & \xmark & \cmark & \cmark \\
\textbf{OAT (ours)}                         & \cmark & \cmark & \cmark & \cmark & \cmark \\
\bottomrule
\end{tabular}
\caption{Structural comparison of OAT with closely related models.
         \cmark~= present; \xmark~= absent; $\sim$~= partial.}
\label{tab:comparison}
\end{table}

\section{Methodology}
\label{sec:methodology}

\subsection{Problem Formulation}

\begin{figure}[h]
    \centering
    \includegraphics[width=\textwidth]{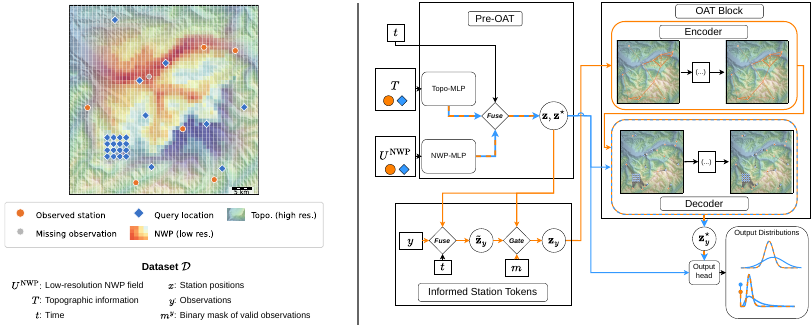}
    \caption{{\bf Schematic of data and observation assimilation transformer (OAT).} Left: Data snapshot with high-resolution topography, coarse NWP field, station locations (orange circles) and illustrative query locations (blue diamonds). Query locations can lie on a grid or at arbitrary locations. Right: Schematic of the OAT architecture. Topography and NWP fields are first embedded at observed and query locations in the {\it Pre-OAT stage}. Then observed embeddings are updated with observation values resulting in the {\it informed station tokens}. The {\it OAT block} comprises a self-attention encoder on the observed embeddings, and a cross-attention decoder between the observed embeddings and query embeddings. The query embeddings, informed by the observational context, are then producing a probabilistic prediction via an output head to produce the final {\it output distributions}. The same shared output head can provide predictions on query tokens before going through the OAT block.}
    \label{fig:oat}
\end{figure}

At a given time $t$ we have a low resolution NWP field $U^{\rm NWP} \in \mathbb{R}^{N_U\times C_U}$, where $C_U$ are the variables of the field and $N_U$ is the number of coordinates, where the field is given. Note that we explicitly allow for irregular non-gridded fields. Positions and observations for the $N$ stations in the dataset are denoted by $\boldx^c\in\mathbb{R}^{N\times2}$ and $\boldy^c\in\mathbb{R}^{N \times C_y}$, respectively. We allow for missing values in the observations, either at the channel (variable) level, or at the station level, where the entire observation vector is unavailable. We define a binary mask $\mathbf{m}^y\in\{0,1\}^{N\times C_y}$, where 1 and 0 denote a valid and invalid observation, respectively. Additionally, we have high-resolution topographic data $T\in\mathbb{R}^{N_T\times C_T}$, which as for the NWP can be an irregular field. We define all the given information in the set $\mathcal{D}_t$ at a given time $t$. The aim is to predict $p_\phi(\boldy^\star_t\vert \boldx^\star, \mathcal{D}_t)$, an informed distribution over the target variables $\boldy^\star$ given \emph{any} set of points $\boldx^\star$ on the considered domain. The distribution is parametrized by $\phi$. An example snapshot of the data can be seen in Figure~\ref{fig:oat} in the left panel, while the right panel illustrates the general architecture of OAT.

\subsection{Observation Assimilation Transformer} \label{subsec:method-oat}

\paragraph{Predictions without observations}
First, we construct a model, akin to classical postprocessing~\citep{rasp2018neural}, that allows predicting anywhere without observations, i.e. $p(\boldy^\star_t\vert \boldx^\star, \mathcal{D}_t\backslash\{\boldy_t^c,\boldx^c\})$ (see Pre-OAT block in Figure~\ref{fig:oat}). We do that for any point of interest $i$ at position $\boldx^\star_i$. We construct a joint non-linear embedding $\boldz_{t,i}=e(U^{\mathrm{NWP}}_{t,i}, T_i, t)$ containing all the information we have at any location $i$ and time $t$. $U^{\mathrm{NWP}}_{t,i}$ and $T_i$ are obtained by interpolation at the points of interest given the field information. Both are embedded with separate multi-layer perceptrons (MLP), and then conditionally fused to a joint embedding $\boldz_{t,i}\in \mathbb{R}^{\dmodel}$. To this end, we use adaptive layer norm, where the shift- and scale-parameters are conditioned on the features extracted from time $t$, i.e. sine and cosine embeddings of day of the year and time of the day (see Appendix~\ref{app:arch-details}). $\dmodel$ is a crucial hyperparameter and indicates the dimensionality of the latent space. Note that we do not provide the absolute coordinates $\boldx^\star$ themselves such that the model can also be used outside of the training domain and does not overfit to coordinate particularities. Finally, we can use an output head $\phi_{t,i}=o(\boldz_{t,i})$ mapping to the parameters of the distributions of the target variables, allowing us to make predictions {\it without}~station observations $\boldy_t$. We call this distribution without any information from station observations \emph{pre-OAT distribution}.

\paragraph{Observation assimilation at station locations}
With OAT, we aim at creating an observation-informed embedding at the station locations $\boldx^c$, where we have observations $\boldy^c$. To this end, we simply fuse the latent embedding $\boldz$ with the observations $\boldy^c$ via another adaptive layer norm.



The fusion layer $\tilde\boldz_y = {\rm Fuse}(\boldz, \boldy, t)$, conditioned on time $t$, incorporates the observation into a preliminary \emph{informed} embedding. If an observation is missing, the observation is set to $0$. To control the information flow based on which values are observed, we design a learnable \emph{gate} that depends on the mask of observed values $\mathbf{m}^y_t$; entries are set to 1 for each location and variable that was observed and 0 otherwise. This leads to
\begin{equation}
    \boldz_y = (1-g(\mathbf{m}^y))\boldz+g(\mathbf{m}^y)\tilde\boldz_y, 
\end{equation}
where $g:\{0,1\}^{N\times C_y}\rightarrow [0,1]^{N\times \dmodel}$ is a learnable MLP. This way, the gate is able to funnel the information flow per dimension of the embeddings. 

\paragraph{Observation assimilation via Transformer}
Next, we want to propagate the information obtained at the station locations $\boldx^c$ to our prediction locations $\boldx^\star$. At the former, we have the informed embedding $\boldz_y$, while at the latter the uninformed ones $\boldz^\star$. We use a customized transformer, where we have queries $\boldz^\star$ and keys and values $\boldz_y$ as input. Compared to the standard encoder-decoder transformer~\citep{vaswani2017attention}, we remove the self-attention among queries in the decoder blocks to avoid that other queries influence the prediction at a given position. Also, positional encoding is discarded. Instead, in the spirit of the translation equivariant transformer, the attention logits between two locations $i,j$ are given by
\begin{equation}
    \alpha_{i,j} = 
    \begin{cases}
    Q_i^\top K_j + \bAttn(\boldx_i^{\rm elev} - \boldx_j^{\rm elev}) & \text{ if } j\in \mathcal{N}_i, \\
    -\infty & \text{ otherwise },
    \end{cases}
    \label{eq:attention}
\end{equation}
where $Q=W_Q\boldz^\star$ are the queries, $K=W_K\boldz_y$ are the keys, $\mathcal{N}_i$ is a neighborhood set for $i$. $\boldx^{\rm elev}$ denotes the location concatenated with the elevation at this position. This attention mechanism differs from standard attention in two ways: First, the attention bias depends on the signed distance between locations, allowing the model to incorporate distance and direction information in the attention. Second, sparse attention allows us to define a graph between stations in the transformer encoder, and stations as well as prediction points in the decoder, respectively. This allows for scalability as the number of stations grows, and the receptive field can be enlarged by adding layers. How to define the neighborhood sets and therefore the graph is a design choice. Simple options include k-nearest neighbor construction or a radius graph, though more sophisticated approaches such as terrain-informed graphs are also possible. For details and alternative attention mechanisms see Appendix~\ref{app:arch-details}.

\paragraph{Informed predictions anywhere}
Since OAT allows us to have informed embeddings $\boldz_y^\star$ anywhere, predictions at arbitrary locations, such as on a high-resolution grid, become straightforward. We use the \emph{same} output head that we used already for the pre-OAT output to have the final distribution parameters $\boldsymbol{\phi}_y^\star=o(\boldz_y^\star)$, which yields the desired distribution $p_{\phi(\boldz_y^\star)}(\boldy^\star_t\vert \boldx^\star, \mathcal{D}_t)$.

\paragraph{Model training} 

At each training step, the observed locations $\boldx^c$ are randomly partitioned into a conditioning set, whose observations are provided to the model, and a holdout set, whose observations are withheld and used only as prediction targets to supervise the model. We use the continuous ranked probability score (CRPS) on each observed variable as the loss function. To encourage best possible predictions even without observational data, we also add a loss term for the pre-OAT predictions. The total loss is given by
\begin{equation}
    \ell(\boldy^\star) = (1-w_\ell)\ell_{\phi_y^\star}(\boldy^\star) + w_\ell\ell_{\phi^\star}(\boldy^\star),
\end{equation}
which is differentiable with respect to all model parameters, where $\boldy^\star$ is given during training. We set the weight $w_\ell=0.1$ throughout this work. Pseudo-code of the training procedure is provided in Algorithm~\ref{alg:training}. For further implementation details, refer to Appendix \ref{app:arch-details} and Table~\ref{tab:training-hyperparams}.

\noindent
\begin{minipage}[t]{\textwidth}
\begin{algorithm}[H]
\caption{OAT Training Step}
\label{alg:training}
\KwIn{NWP fields $U_t^{\rm NWP}$, stations $(\boldx, \boldy_t)$, topography $T$}
$(\boldx^{c},\boldy^{c}),\ (\boldx^{*},\ \boldy^{*}) \leftarrow \mathrm{RandomSplit}((\boldx,\boldy))$\;\tcp{Random 80/20 partition of stations}

\For{each location $i$ in $\boldx^{c} \cup \boldx^{*}$}{
    $\boldz_{i} \leftarrow e(U_{t,i}^{\rm NWP},\ T_i,\ t)$\; \tcp{Embed and fuse NWP prediction and topography cond. on time}
}
$\phi^{*} \leftarrow o(\boldz^{*})$\;\tcp{Get pre-oat predictive distribution}

\For{each station $i$ in $\boldx^{c}$}{
	$\tilde{\boldz}_{i} \leftarrow {\rm Fuse}(\boldz_i, \boldy_{i}, t)$\;
	\tcp{Incorporate observations}
	$\boldz_{y,i} \leftarrow (1 - g(\mathbf{m}^{y}_i))\,\boldz_{i} + g(\mathbf{m}^{y}_i)\,\tilde{\boldz}_{i}$\;\tcp{Gate for (partial) observation}   
	}

$\boldz^{*}_{y} \leftarrow {\rm Transformer}(\boldz^{*},\ \boldz_{y})$\;\tcp{Propagate via transformer}

$\phi^{*}_{y} \leftarrow o(\boldz^{*}_{y})$\;\tcp{Get oat predictive distribution}
$\ell \leftarrow (1-w_\ell)\ell_{\phi^{*}_{y}}(\boldy^{*}) + w_\ell\ \ell_{\phi^{*}}(\boldy^{*})$\; \tcp{Compute loss (and gradients)}

\end{algorithm}
\end{minipage}

\section{Experiments}
\label{sec:experiments}

\subsection{Data}
\label{sec:datasets}

\begin{figure}
    \centering
    \includegraphics[width=1\linewidth]{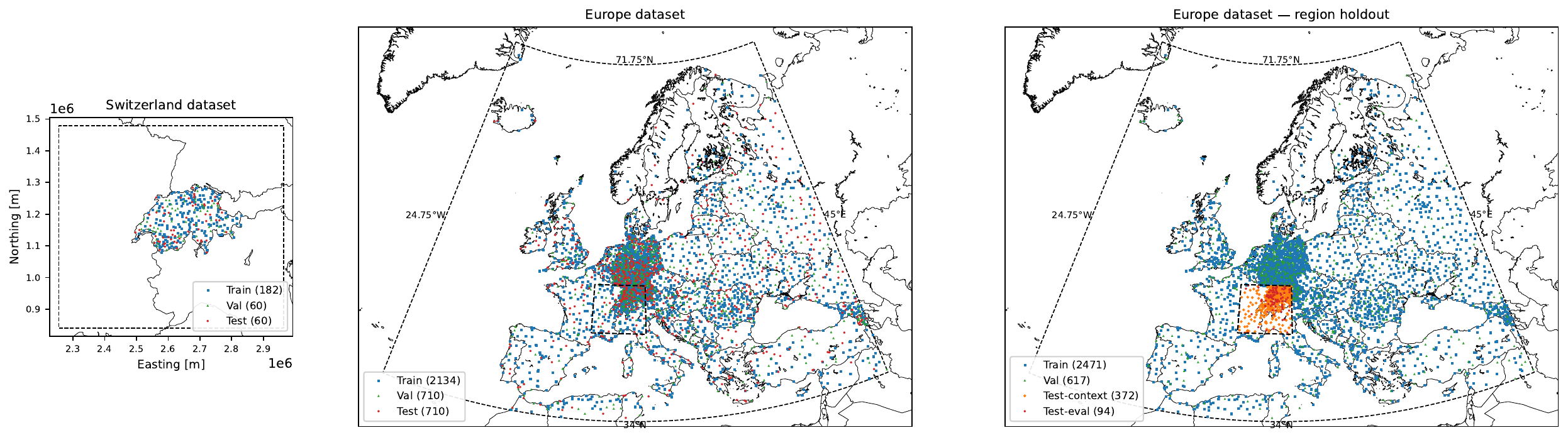}
    \caption{{\bf Dataset extent and station distribution} The dashed line shows the extent for which the NWP product is available. The square, triangle, and circle markers are train, validation, and test stations, respectively. Left: The Switzerland dataset. Center: Europe dataset. Right: Same data as the Europe dataset, but with a regional hold-out split, which coincides with small dashed box on the right representing the extent Switzerland dataset.}
    \label{fig:data}
\end{figure}

We train and evaluate our model on two datasets that differ in spatial scale, NWP source, and station network, allowing us to assess generalization in different settings. Both datasets span $\sim$5-6 years at hourly resolution and share the same three-part structure: an NWP deterministic prior, high resolution topography with additional features derived from it, and ground truth sparse observations. We perform experiments at two spatial scales, covering a regional scale around Switzerland and a broader European scale (see Figure~\ref{fig:data}).

\paragraph{Switzerland dataset} For the first dataset we consider Switzerland, which poses a difficult challenge due to its complex topography. We take the COSMO-1 operational analysis archive of MeteoSwiss (see~\citet{zanetta2026efficient} for details) as NWP input field, covering Switzerland at a horizontal grid spacing of \SI{\sim1.1}{\kilo\metre} (0.01\textdegree). It is a regional forecasting system operated by MeteoSwiss, and its hourly analysis is obtained via an ensemble Kalman filter data assimilation scheme. We use four physical fields that are considered to be directly relevant for the later targets (temperature and precipitation): \SI{2}{\metre} temperature ($T_\mathrm{2m}$), ground temperature ($T_G$), hourly total precipitation, and total cloud cover fraction.

For each station and query location, terrain features are sampled from a high-resolution digital elevation model (DEM) at \SI{\sim50}{\metre} resolution. Besides the elevation itself, four terrain descriptors are used: topographic position index (TPI) at \SI{500}{\metre} and \SI{2000}{\metre} scales~\citep{dereu2013application}, and
west--east and south--north terrain slope derivatives at \SI{2000}{\metre} scale. These descriptors are precomputed on the DEM grid, so multi-scale terrain context enters through the descriptors themselves rather than through a spatial patch. 

Surface observations are sourced from the PeakWeather dataset~\citep{zambon2025peakweather}, which provides access to measurements from the SwissMetNet, the automatic station network of MeteoSwiss. The measurements are available at 10-minute resolution and are aggregated to hourly for this work. For temperature the aggregation function is the average, and for precipitation it is the sum. The network comprises 302 stations across Switzerland.
Two target variables are considered: 2-m temperature (\si{\celsius}) and hourly precipitation accumulation (\si{\milli\metre\per\hour}). Station coordinates are expressed in the Swiss national reference frame (CH1903+/LV95). The data span 2019-07-03 to 2024-09-04.

\paragraph{Europe dataset}
The second dataset has a continental scale, where we consider the spatial extent of the European continent and surroundings ($34\text{--}71.75^{\circ}\mathrm{N}$, $24.75^{\circ}\mathrm{W}\text{--}45^{\circ}\mathrm{E}$). The NWP field consists of ERA5 reanalysis fields \citep{https://doi.org/10.1002/qj.3803} obtained from ECMWF via the MARS archive on the native N320 reduced Gaussian grid, whose spacing is \SI{\sim31}{\kilo\metre}. It is a global atmospheric reanalysis produced by ECMWF, generated by assimilating observational data into the Integrated Forecast System (IFS) using 4D-Var \citep{https://doi.org/10.1111/j.1600-0870.1986.tb00459.x}. Again, three fields are used that are directly relevant for the target: 2-m temperature, total precipitation, and total cloud cover fraction.

Two static surface descriptors are provided per station: terrain elevation is taken from the Terrain Tiles~\citep{terraintiles} at 3 arc-seconds (\SI{\sim90}{\metre}) and is resampled to \SI{0.01}{\degree}. Additionally, we compute distance to the nearest coastline at the same resolution, which serves as a proxy for the maritime influence on local climate.

Hourly surface observations are acquired from the \href{https://dev.meteostat.net/cli}{Meteostat CLI} and its data providers (Deutscher Wetterdienst for Germany, GeoSphere Austria for Austria, and NOAA ISD, License: CC BY 4.0), a consolidated open-access database of global surface stations. After quality filtering, the network comprises 3554 stations distributed across Europe. Two target variables are considered: 2-m temperature (\si{\celsius}) and hourly precipitation accumulation (\si{\milli\metre\per\hour}). In Figure~\ref{fig:data} we see that some regions are more densely sampled than others (e.g. Germany has a much denser sampling rate). This is due to station availability in the Meteostat interface. The archive spans 2019-01-01 to 2024-12-31.

\paragraph{Europe dataset with region hold-out} The third dataset is the same as the second one, except that an entire region encompassing Switzerland and its neighboring areas is held out during training. This configuration enables an assessment of the spatial generalization ability of the tested models to unobserved regions.
At test time, $80\%$ of the hold-out stations are selected as conditioning stations, while the remaining $20\%$ are reserved as final test stations.

\paragraph{NWP inperpolation and local readout}
Both NWP archives are delivered as unstructured lists of grid points. The subsequent methodology requires NWP fields at arbitrary locations. Hence, we interpolate the NWP fields with Delaunay triangulation, since NWP grids are irregular in both datasets. The topographic variables are provided at relatively high resolution and are difficult to interpolate meaningfully (e.g. TPI). Hence, we resort to just take the value at the nearest available location. 

While the Swiss coordinate system allows the use of Euclidean distance metrics with negligible distortion within the Swiss domain, this does not hold for the Europe dataset. Distances in planar coordinates become increasingly distorted toward higher latitudes. We therefore project coordinates onto the sphere and compute Euclidean distances in the three-dimensional space. Consequently, $\boldx^{\rm elev}$ is four dimensional with coordinates in 3D space and elevation. Since we only consider a nearest-neighbor network, this is a fair approximation.

For a valid training procedure, we use an identical, dataset-agnostic splitting procedure for both datasets, applied separately along the temporal and station (spatial) axes (Figure~\ref{fig:data} left and center).
Stations are randomly permuted and partitioned into disjoint train/validation/test sets with a nominal 60\%/20\%/20\% split, never shared across splits. This yields 182/60/60 stations (of 302) for the Swiss dataset, and 2134/710/710 stations (of 3554) for the Europe dataset. The splits are the same across all models. A single contiguous temporal cutoff reserves the last 365 days of each dataset as the test period; the remaining, earlier period is shared by train and validation alike. It should be noted that train and validation are therefore not separated in time, only test is temporally held out -- validation is a purely spatial (station) hold-out to make use of most of the data. For the Switzerland dataset, train/val covers 2019-07-03 to 2023-09-05, and test covers 2023-09-06 to 2024-09-04; in the Europe dataset, train/val covers 2019-01-01 to 2024-01-01, and test covers 2024-01-02 to 2024-12-31 (365 days). Although training and validation periods should ideally be disjoint, we allowed them to overlap in order to make most efficient use of the data. This choice does not compromise the validity of the reported results, as all evaluations are conducted on a separate, non-overlapping test period.

Since OAT does not rely on absolute coordinates, it could, in principle, also make predictions in areas where it was not trained. To test this, we use the Europe dataset to construct a spatial hold-out experiment, where all stations in the Swiss domain are excluded from the training set (Figure~\ref{fig:data} right). This yields 2471 training stations and 617 validation stations. Then we define 80\% of the stations (372) within the test window as context, to predict the remaining 20\% (94). This split also allows us to directly benchmark OAT against COSMO fields. The training period is 2019-01-01 to 2024-01-01. The test period is 2024-01-02 to 2024-09-04, as COSMO is not available for the full 2024 year.

\subsection{Baselines} We consider five different baselines. The first is the {\it NWP interpolator}, which does not involve learning parameters but uses interpolation of the NWP field of the target variable, and takes that as prediction. The second is the {\it uninformed baseline}, which uses only the first part of OAT, using the uninformed embedding for predictions, i.e. there is no observation assimilation transformer block. This makes the uninformed baseline a feedforward neural network, trained on weather stations, that tries to predict at test locations based on the interpolated NWP field, the topography, and the time-embedding, but without the ability of data assimilation. As a third baseline, we use the Gaussian process methodology by~\citet{zanetta2026efficient} in two variants. In both the prior mean is predicted by an MLP, as in the uninformed baseline. In the {\it GP} baseline, we use an automatic relevance determination (ARD) kernel as the covariance function. In the {\it  DK-GP} (deep kernel GP), the covariance function is an ARD kernel multiplied by a deep RBF kernel, following~\citet{zanetta2026efficient} (see Appendix~\ref{app:gp} for details). Since the GP framework requires a Gaussian likelihood for exact inference, we train this model on the log-transformed (with $10^{-3}$ offset) precipitation values. However, we track the CRPS in the original data space on the validation set for model selection. Furthermore, while it is possible in principle, no multivariate code for GP is provided, hence we train every variable separately. As an additional baseline, we consider the gridded TNP~\citep{ashman2025gridded}, encoding NWP and observations on a latent pseudo-token grid ($80\times120$ for Switzerland, $40\times72$ for Europe; see Appendix~\ref{app:gridded-tnp}), where then Swin attention is used to encode the global information. The decoder then provides the predictive distribution at the points of interest. For fairness, while different from the original model, we provide the decoder with the high-resolution topographic data. Like the OAT, all models are trained with the closed form CRPS loss of each predictive distribution (see Appendix \ref{app:losses}). We use the AdamW optimizer \citep{loshchilov2018decoupled} with cosine annealing learning rate scheduling for all models except the Gridded TNP. For the latter, we used Adam and no scheduler. The hyperparameters used for OAT are given in Table~\ref{tab:oat-hyperparams} in the Appendix.

\begin{figure}
    \centering
    \includegraphics[width=1\linewidth]{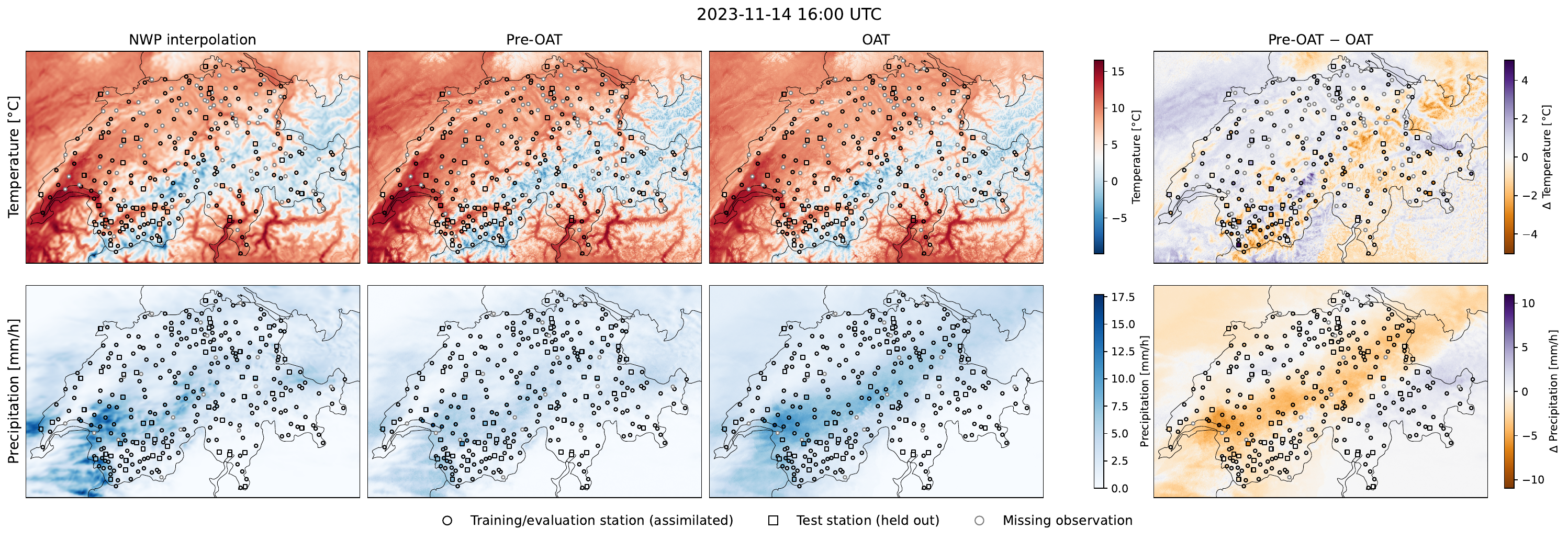}
    \caption{{\bf Example of predicted fields} From left to right: The interpolated NWP, the pre-OAT predicted, and the OAT field with assimilated and test stations. The color code of the stations are the measurement values. Rightmost is the difference between the pre-OAT and OAT field, where the station color code is between pre-OAT and measurement at that station. Top row: Temperature fields. Bottom row: Precipitation fields. The prediction fields are sampled on a regular \SI{250}{\metre} grid.}
    \label{fig:example_field}
\end{figure}

\subsection{Predictive performance of OAT}
In Figure~\ref{fig:example_field} we show illustrative predictive (mean) fields of temperature and precipitation in the Switzerland dataset queried at 250m resolution. To see the effect of the observation assimilation, we make use of OAT's design, and compare the predictive field of its pre-OAT distribution (no observation assimilated) with its final prediction. The temperature fields of both the prior and final OAT are more finely resolved compared to the NWP, demonstrating that both successfully incorporate the high-resolution topography information. The difference between the two fields indicates that the observation assimilation transformer has a correcting effect on the fields by moving the fields closer to the observed station temperatures.
While topography seems to play a less important role for precipitation, the observation assimilation effect is even stronger. While the pre-OAT field follows closely the NWP field, the OAT predicts a wide precipitation band along the north of the Alps, closely matching the observed station values at assimilated and test locations. We would like to emphasize that the figures show the mean of the predictive distributions; these are therefore expected to appear smoother than physical precipitation patterns.

\begin{table}[h]
  \centering
  \begin{tabular}{ll|ccc|ccc}
    \toprule
    & & \multicolumn{3}{c|}{\textbf{Temperature}} & \multicolumn{3}{c}{\textbf{Precipitation}} \\
    \cmidrule(lr){3-5}\cmidrule(lr){6-8}
    \textbf{Dataset} & \textbf{Model} & MAE $\downarrow$ & CRPS $\downarrow$ & Skill $\uparrow$& MAE $\downarrow$ & CRPS $\downarrow$ & Skill $\uparrow$\\
    \midrule
    \multirow{6}{*}{\textbf{Switzerland}} & COSMO Interp.  & $1.287$ & $1.287$ & $-0.637$ & $0.146$ & $0.146$ & $-0.587$ \\
    & Uninf. baseline  & $1.089$ & $0.786$ & $0.000$ & $0.122$ & $0.092$ & $0.000$ \\
    & GP  & $1.109$ & $0.815$ & $-0.037$ & $0.130$ & $0.111$ & $-0.206$ \\
    & DK-GP & $1.125$ & $0.827$ & $-0.052$ & $0.119$ & $0.101$ & $-0.094$ \\
    & Gridded TNP & $1.429$ & $1.115$ & $-0.418$ & $0.097$ & $0.073$ & $0.207$ \\
    & OAT & $1.046$ & $0.751$ & $0.045$ & $0.086$ & $0.064$ & $0.306$ \\
    \midrule
    \multirow{6}{*}{\textbf{Europe}} & ERA5 Interp. & $1.305$ & $1.305$ & $-0.445$ & $0.106$ & $0.106$ & $-0.623$ \\
    & Uninf. baseline & $1.247$ & $0.903$ & $0.000$ & $0.082$ & $0.065$ & $0.000$ \\
    & GP & $1.189$ & $0.934$ & $-0.036$ & $0.086$ & $0.076$ & $-0.170$ \\
    & DK-GP & $1.145$ & $0.874$ & $0.031$ & $0.083$ & $0.074$ & $-0.134$ \\
    & Gridded TNP & $1.178$ & $0.882$ & $0.023$ & $0.077$ & $0.060$ & $0.072$ \\
    & OAT & $0.992$ & $0.714$ & $0.209$ & $0.067$ & $0.052$ & $0.203$ \\
    \midrule
    \multirow{4}{*}{\shortstack[l]{\textbf{Europe}\\(region hold-out)}} & ERA5 Interp. & $1.702$ & $1.702$ & $-0.376$ & $0.160$ & $0.160$ & $-0.668$ \\
    & COSMO Interp. & $1.168$ & $1.168$ & $0.056$ & $0.141$ & $0.141$ & $-0.471$ \\
    & Uninf. baseline & $1.701$ & $1.237$ & $0.000$ & $0.118$ & $0.096$ & $0.000$ \\
    & OAT & $1.086$ & $0.786$ & $0.365$ & $0.090$ & $0.069$ & $0.277$ \\
    \bottomrule
  \end{tabular}
\caption{{\bf Model performance on the test set.} Skill is computed with CRPS and the comparison is to the uninformed baseline.}
\label{tab:ml_results}
\end{table}

Table~\ref{tab:ml_results} shows the aggregate performance of different models across the test-sets (see Table~\ref{tab:ci_table} for confidence intervals). We compute the mean absolute error (MAE) between the median prediction and the observations, the CRPS, and the skill in CRPS compared to the uninformed baseline on the test stations. In terms of skill, OAT significantly outperforms the uninformed baseline on both datasets and for both variables, except for temperature on the Switzerland dataset, where the difference is not statistically significant. OAT achieves the lowest MAE for temperature, reducing it close to $\SI{\sim 1}{\celsius}$. Surprisingly, on the Swiss domain, the remaining baselines suffer performance degradation compared to the uninformed baseline for temperature. A plausible explanation is that, in such complex topography, most of the relevant information is already contained in the NWP temperature fields and elevation. Hence, the baselines seem to struggle to effectively assimilate station observations in this setting. In particular, the Gridded TNP's performance suffers on the Switzerland dataset. We hypothesize that some information is lost by passing it through the latent $80\times120$ grid, which is relatively coarse and cannot resolve the fine grained topography of Switzerland. For precipitation, by contrast, the Gridded TNP generally improves upon the uninformed baseline, although it is still outperformed by OAT. Moreover, OAT's predictive distributions achieve a lower CRPS. Table~\ref{tab:att_oat_perf} compares alternative attention mechanisms for OAT on the Swiss dataset. The proposed mechanism  (Equation~\eqref{eq:attention}) performs comparably, with only marginal differences between the variants.

\begin{figure}[h]
     \centering
     \begin{subfigure}[b]{0.49\textwidth}
         \centering
         \includegraphics[width=1.0\linewidth]{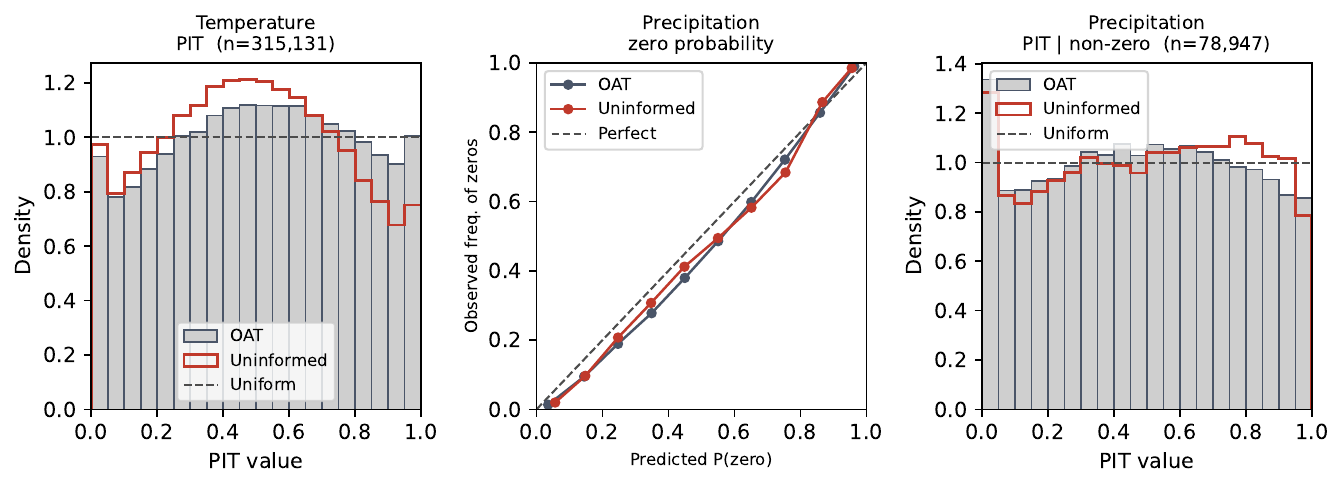}
         \caption{Calibration on Switzerland dataset.}
     \end{subfigure}
     \hfill
     \begin{subfigure}[b]{0.49\textwidth}
         \centering
         \includegraphics[width=1.0\linewidth]{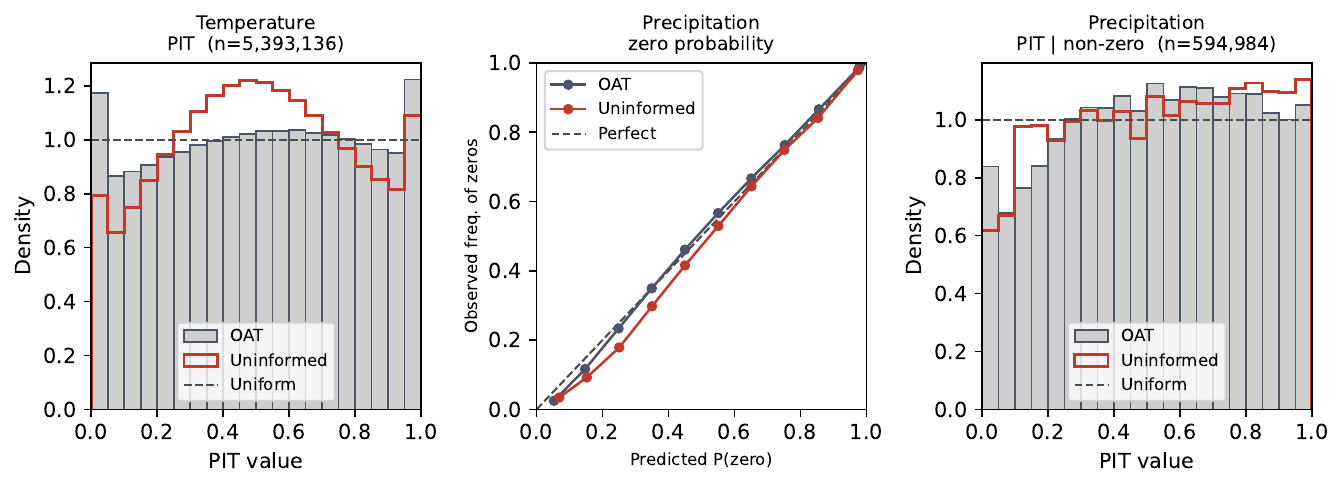}
         \caption{Calibration on Europe dataset}
     \end{subfigure}
    \caption{{\bf Calibration for the OAT on the test set.} In each panel, from left to right: the PIT histogram for temperature, the reliability diagram for the probability of zero precipitation, and the precipitation PIT histogram for wet events.}
    \label{fig:pit}
\end{figure}

We assess probabilistic calibration with the sample-based probability integral transform (PIT) histogram. For the zero-inflated distribution of precipitation, we split the analysis into a reliability diagram for the dry events and a PIT histogram for the wet events (see Appendix~\ref{app:calibration}). The calibration of the uninformed baseline and the OAT model for both datasets is presented in Figure \ref{fig:pit}. For Switzerland, we see that both models suffer from overdispersion for temperature, i.e. their predictive distributions are too conservative, although OAT shows slight improvement over the uninformed baseline. For precipitation, the uninformed baseline is even marginally better calibrated. We hypothesize that this is due to the comparatively small number of stations in the Switzerland dataset, as the picture changes for the Europe dataset, where OAT is better calibrated for both temperature and precipitation.

\begin{figure}[h]
 
    \begin{subfigure}[b]{0.49\textwidth}
         \centering
        \includegraphics[width=1\linewidth]{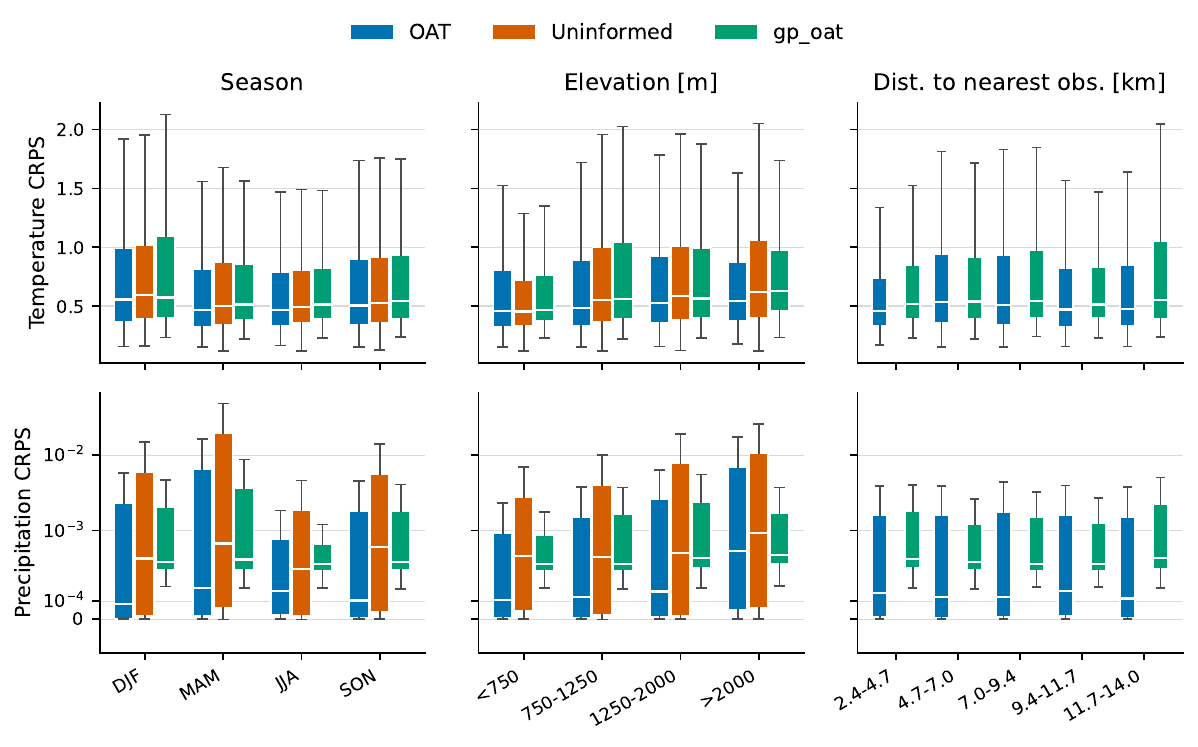}
        \caption{Switzerland dataset}
        \label{fig:error_cosmo}
    \end{subfigure}
    \begin{subfigure}[b]{0.49\textwidth}
        \centering
        \includegraphics[width=1\linewidth]{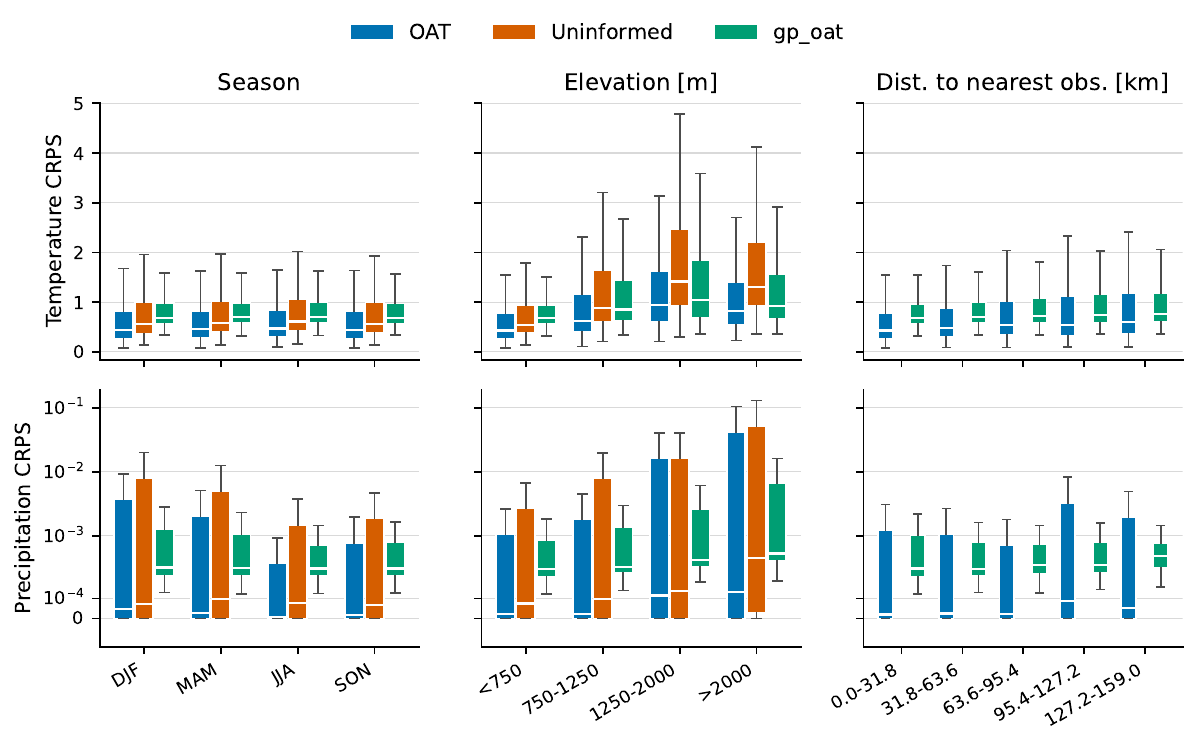}
        \caption{Europe dataset}
        \label{fig:error_era5}
    \end{subfigure}
    \caption{{\bf Stratification analysis of CRPS} From left to right, the CRPS on the test set as function of season, elevation, and distance to closest observed station. The white bar denotes the median, and the boxes are the first and third quartile. Since the uninformed baseline is not conditioned on any observation, it is not displayed in the distance panels.}
\end{figure}

In order to understand the performance of OAT better, we stratify the CRPS in Figure~\ref{fig:error_cosmo} and~\ref{fig:error_era5}. For seasonality, the temperature CRPS stays relatively constant, while precipitation during the summer months shows less variable CRPS. For precipitation OAT performs best across all seasons. With respect to elevation, the CRPS increases with altitude for both variables and all models, with OAT performing best across the entire range. On the Switzerland dataset, this increase is particularly pronounced for precipitation at high altitudes ($>$\SI{2}{\kilo\metre}). We further examine how the distance to the nearest observed station affects the CRPS. This is of particular interest for the Europe dataset, where station density varies considerably. As shown in Figure~\ref{fig:error_era5}, the CRPS increases slightly with distance for temperature. For precipitation, it remains relatively stable up to a distance of $\SI{\sim100}{\kilo\metre}$, beyond which it rises noticeably. This suggests a critical distance beyond which an observation ceases to be informative.

\begin{figure}[h!]
    \centering
    \includegraphics[width=1\linewidth]{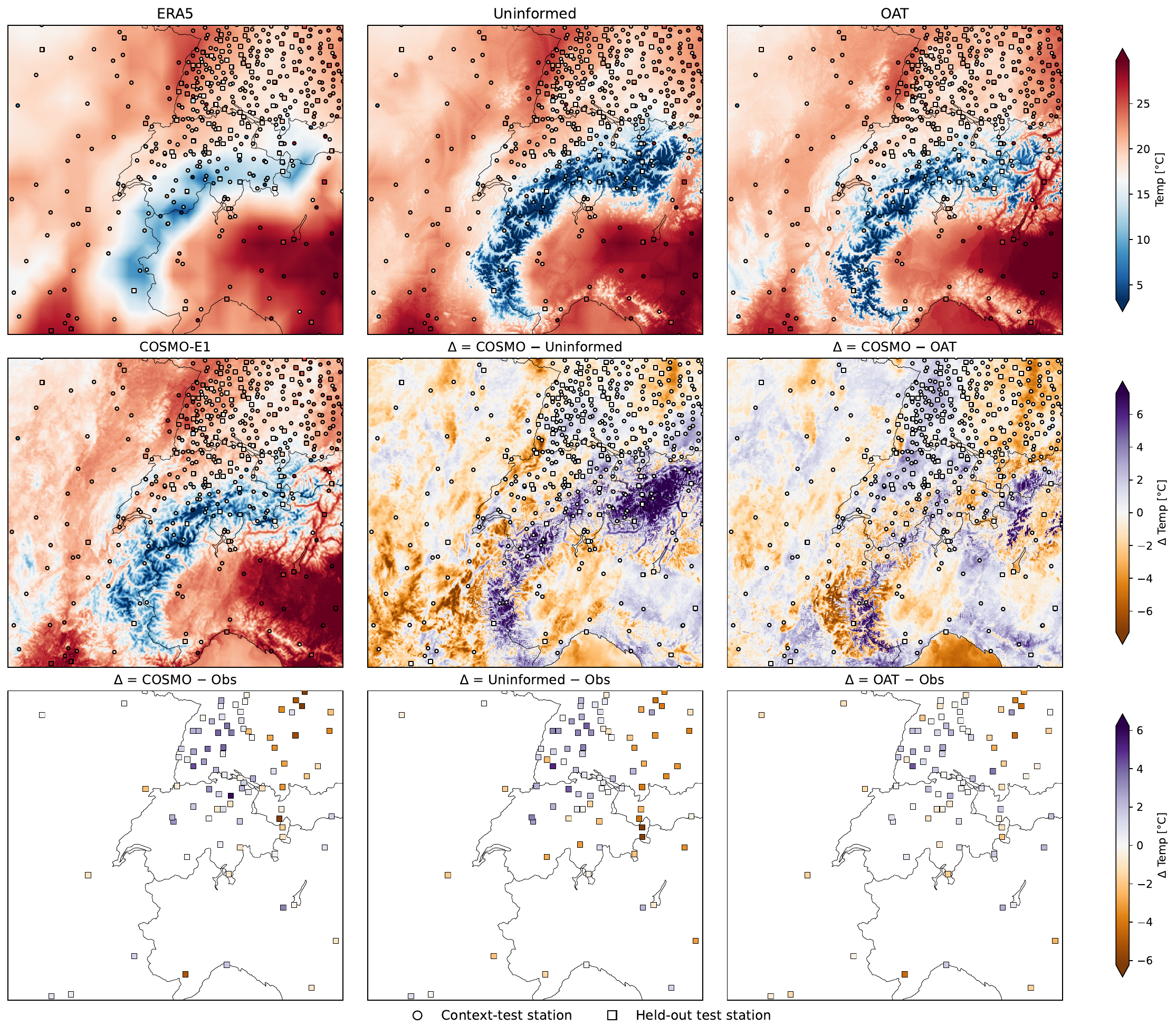}
    \caption{{\bf Region hold-out and downscaling temperature} from the ERA5 grid to the COSMO grid ($\sim$1 million points) on 2024-06-21 12:00 UTC for Swiss domain. In the first row, the input ERA5 field, the uninformed mean prediction, and the OAT mean prediction. In the second row, the corresponding COSMO analysis field as well as the differences between the informed and uninformed models against that analysis. The third row shows the test stations and the corresponding prediction error compared to the observed temperature for COSMO, the uninformed baseline, and OAT.}
    \label{fig:era5_to_cosmo}
\end{figure}

\subsection{Region hold-out and downscaling from ERA5 to COSMO}

One important design choice of OAT was to not provide the model with the absolute positions of stations or query points, but only with distance information. In principle, this would allow for spatial transferability, i.e. making predictions in a region on which OAT was not trained. To investigate whether OAT's good predictive performance persists in such a scenario, we train on the Europe dataset excluding all stations within and around Switzerland (Figure~\ref{fig:data} right). This setup also allows assessment of OAT's capabilities against the high resolution analysis product COSMO (\SI{\sim1}{\kilo\metre}), while only providing OAT the NWP information from the ERA5 N320 reduced Gaussian grid (\SI{\sim31}{\kilo\metre}). We provide 80\% of the stations in the hold-out domain as condition and predict on the remaining 20\%. The results are provided in the bottom of Table~\ref{tab:ml_results}. OAT slightly outperforms COSMO for temperature in MAE, even though COSMO might have had access to some of the test stations. Furthermore, OAT improves the ERA5 interpolation, establishing transfer capabilities in this setting. For a particular example of downscaling the temperature field and comparison to COSMO see Figure~\ref{fig:era5_to_cosmo}. OAT also outperforms COSMO interpolation for precipitation.

\begin{figure}[h]
    \centering
    \includegraphics[width=1\linewidth]{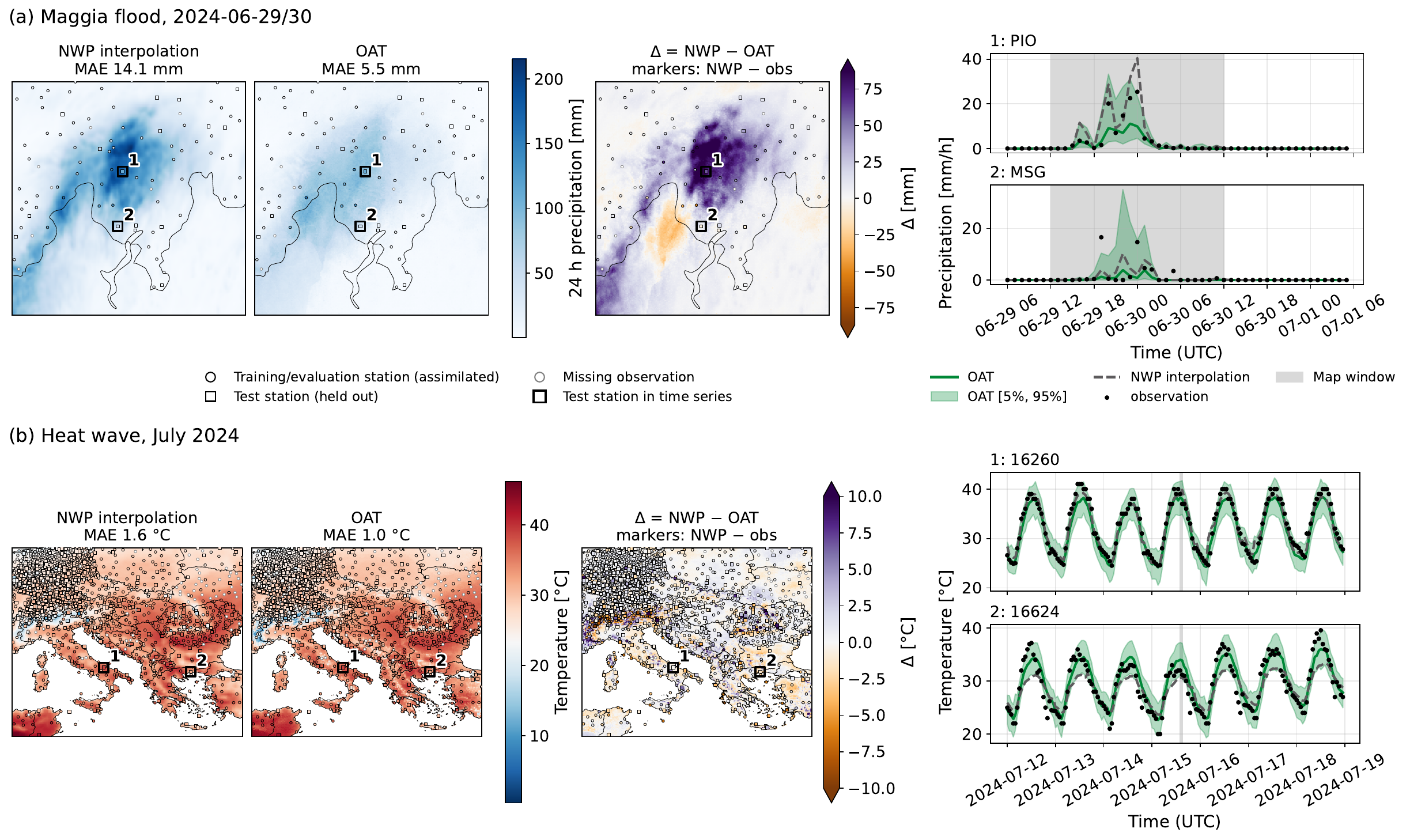}
    \caption{{\bf  OAT's predictive performance for exemplary extreme events.} Top: Extreme precipitation event causing flooding in the Maggia valley end of June 2024. Left: Cumulated precipitation map over 24h for NWP, OAT and their difference, respectively. Right: Time-series of precipitation at two test-stations. Bottom: Heat wave in July 2024 in south eastern Europe. Left: Temperature maps of NWP (COSMO/ERA5) and OAT, with differences. Right: Two test-stations for one week.}
    \label{fig:extreme_events}
\end{figure}

\subsection{Case study: Predictions on extreme events}
A crucial capability for data-driven models is to perform well during extreme events. Hence, we investigate OAT's performance on two events, one for the Switzerland dataset and one for the Europe dataset (without hold-out). Both events are in the corresponding test sets. The results are shown in Figure~\ref{fig:extreme_events}. The first case is the Maggia flood on 2024-06-29/30, triggered by an extreme precipitation event. For the accumulated precipitation over this period, OAT achieves a substantially lower MAE than COSMO at the test stations in the affected region (\SI{5.5}{\milli\metre} vs. \SI{14.1}{\milli\metre}). The field plot shows that COSMO predicts considerably more rainfall than OAT. Given the extreme nature of the precipitation event, accurate predictions at specific stations are inherently difficult. OAT accounts for this by drastically increasing its predictive uncertainty during the event, which is a desirable behavior of a probabilistic model. The second extreme event is the heatwave in southeastern Europe in July 2024. Here, OAT outperforms the ERA5 interpolation in terms of MAE at the test stations (\SI{1.0}{\celsius} vs. \SI{1.6}{\celsius}) for the timestep shown in the lower right panel in Figure~\ref{fig:extreme_events}. Moreover, OAT's predictive distribution closely tracks the observed temperature throughout the week of the event (Figure~\ref{fig:extreme_events}, lower right).

\section{Discussion}
\label{sec:discussion}

Inspired by classical statistical post-processing and recent advances in neural processes~\citep{ashman2024translation,ashman2024approximately}, we introduced OAT, a novel architecture that refines NWP fields by leveraging high-resolution topographic information and sparse station observations. 
Benchmarked against several strong baselines, OAT achieves competitive predictive performance on both a small and a large scale dataset, with the latter demonstrating the scalability of our approach (see Appendix~\ref{app:scalability} for detailed analysis). The model is highly flexible with respect to the choice of output distributions and handles missing observations in a straightforward manner. A particularly useful feature of OAT is its ability to inspect predictions before and after observations are assimilated, making it possible to quantify the contribution of the observational input. We further showed that OAT generalizes to regions not seen during training, enabling spatial transferability, while its performance is limited at the tails of the training distribution (e.g., precipitation at high altitudes). Finally, in case studies of extreme events, OAT appropriately increases its predictive uncertainty, indicating robust uncertainty quantification in these scenarios.

Despite these findings, OAT has some limitations. First, the model only captures the marginal predictive distribution at each location, rather than the correlations and joint distribution across locations. Consequently, samples at different locations are conditionally independent: although the marginals are well calibrated, a field composed of independent samples at each location does not represent a physically meaningful precipitation scenario. Second, OAT does not currently incorporate dense radar observation products, which, from a meteorological perspective, are essential for obtaining spatially coherent precipitation fields. Third, OAT provides no straightforward way to distinguish measurement noise from model uncertainty, a separation that arises naturally in the Bayesian GP framework.

Several directions could address these limitations. Modeling correlated noise with neural processes, as proposed by~\citet{bruinsma2021gaussian}, would be a natural extension of OAT, as would the inclusion of additional dense observational inputs. Additionally, spatial coherence could be encouraged through a multivariate scoring rule such as an energy score, a multivariate generalization of the CRPS, evaluated jointly across locations. Alternatively, OAT's architecture could be embedded within conditional generative denoising diffusion~\citep{ho2020denoising} or flow matching frameworks~\citep{lipman2023flow}.

In this work, we focused on the instantaneous refinement of NWP fields. Incorporating past and future observations is a promising direction for further improving predictions. In OAT, this can be achieved naturally by treating time as an additional coordinate alongside location and elevation. OAT could then even be used for forecasting, trained to predict future observations conditioned on past ones, which would make it directly comparable to the models by~\citet{miralles2025observation} and~\citet{yang2024offgrid}.

Beyond the setting considered here, we believe OAT is applicable to a broader range of problems. One example is the refinement of sea surface temperature or ocean current fields using observations from floating surface drifters~\citep{botvynko2025neural}, a challenging task due to the continuously changing positions of the measurements.

In conclusion, OAT is a versatile model that demonstrated strong performance on the NWP refinement task considered here, offering multiple avenues for future extensions.

\section*{Acknowledgements}
This project was conducted as part of the MeteoSwiss-Swiss Data Science Center collaborative framework agreement. Michele Cattaneo was supported by the EUMETNET Artificial Intelligence and Machine Learning for Weather, Climate and Environmental Applications (E-AI) Programme through the 2025 Research Cooperation Grant.

\bibliographystyle{plainnat}
\bibliography{references}

\setcounter{figure}{0}
\renewcommand{\thefigure}{A.\arabic{figure}}
\setcounter{table}{0}
\renewcommand{\thetable}{A.\arabic{table}}
\newpage
\appendix

\section{Output distributions and loss Functions} \label{app:losses}

Models are trained with the continuous ranked probability score (CRPS), a proper scoring rule which measures how well a predicted probability distribution matches an observed outcome in terms of calibration and sharpness. Many parametric distributions have a closed form definition.

\noindent
For the \textbf{Normal distribution} $\mathcal{N}(\mu,\sigma^2)$ and target $y$:
\begin{equation}
\text{CRPS}_\mathrm{N}\!\left((\mu,\sigma^2),\, y\right) = \sigma \left( z\!\left(2\Phi(z)-1\right) + 2\phi(z) - \frac{1}{\sqrt{\pi}} \right),
\end{equation}
where $z=\frac{y-\mu}{\sigma}$ and $\Phi$, $\phi$ are the standard normal CDF and PDF respectively.

\noindent
For \textbf{Log-Normal distribution} with location parameter $\mu_{\rm LN}$ and scale $\sigma_{\rm LN}$ the density function is
\begin{equation}
    p_{\rm LN}(y\vert \mu_{\rm LN}, \sigma_{\rm LN}) = \frac{1}{y\sigma_{\rm LN}\sqrt{2\pi}}\exp\left(-\frac{\left(\ln y-\mu_{\rm LN}\right)^2}{2\sigma_{\rm LN}^2}\right).
\end{equation}
The CRPS is
\begin{equation}
\text{CRPS}_\mathrm{LN}\!\left((\mu_{\rm LN},\sigma_{\rm LN}),\, y\right) = y \!\left(2\Phi(z_{\rm LN})-1\right) - 2m\left(\Phi(z_{\rm LN}-\sigma_{\rm LN}) - \Phi(-\sigma_{\rm LN}/\sqrt{2})\right), 
\end{equation}
with $z_{\rm LN}=(\ln y - \mu_{\rm LN})/\sigma_{\rm LN}$ and $m=\exp(\mu_{\rm LN} + \sigma_{\rm LN}^2/2)$. While we do not use the log-normal distribution it is an important one to e.g. model wind speed. We provide this in the code.

\noindent
For \textbf{zero-inflated Log-Normal distribution} with $\pi_{\rm ZILN}$ the probability of a random variable being non-zero, and $\mu_{\rm ZILN}$ and scale $\sigma_{\rm ZILN}$ being location and scale parameter of the non-zero density, the density is given by
\begin{equation}
    p_{\rm ZILN}(y\vert\pi_{ZILN}, \mu_{\rm ZILN}, \sigma_{\rm ZILN}) = (1-\pi_{\rm ZILN})\delta(y) + \pi_{\rm ZILN}p_{\rm LN}(y\vert\mu_{\rm ZILN}, \sigma_{\rm ZILN}),
\end{equation}
where $\delta(y)$ is a Dirac delta function. The CRPS is
\begin{align}
\text{CRPS}_\mathrm{ZILN}\!\left((\pi_{\rm ZILN},\mu_{\rm ZILN},\sigma_{\rm ZILN}),\, y\right) = & (1-\pi_{\rm ZILN})y + \pi_{\rm ZILN}\text{CRPS}_\mathrm{LN}\!\left((\mu_{\rm ZILN},\sigma_{\rm ZILN}),\, y\right)\nonumber \\
& + 2 (1-\pi_{\rm ZILN})\pi_{\rm ZILN}m(\Phi(\sigma_{\rm ZILN}/\sqrt{2})-1).
\end{align}

\noindent
Alternatively, a Monte Carlo estimation via predicted samples $\{\hat{y_i}\}_{i=1}^{i=M}$ from a generic distribution exists in the form:

\begin{equation}
    \text{CRPS}\!\left(\{\hat{y_i}\}_{i=1}^{M},\, y\right) = \frac{1}{M}\sum_{i=1}^M \mid \hat{y}_i - y \mid- \frac{1}{2M(M-1)}\sum_{i=1}^M \sum_{k=1}^M \mid \hat{y}_i - \hat{y}_k \mid
\end{equation}
We use the closed form losses in normalized space for training and the MC estimate for validation metrics using denormalized samples.

\section{OAT Architectural Details} \label{app:arch-details}

\subsection*{Fusion layer:} The fusion layer ${\rm Fuse}(a, b, c)$ combines information from $a$ and $b$ into a single vector conditioned on $c$. First, $a$ and $b$ are concatenated as $x = [a\|b] \in \mathbb{R}^d$ and then passed through a series of blocks as follows:

%

\begin{equation}
\begin{aligned}
\tilde{x} & = \text{AdaLN}(x^l, c) \\
x^{l+1} & = x^l + \text{Dropout}_p \left( \text{GLU}(W\tilde{x})\right) \\
\end{aligned}
\end{equation}
where $\text{GLU}(w) = u \otimes \sigma(v)$ is a gated linear unit function, where $ w = [u\| v]$, i.e $u$ and $v$ split $w$ in two halves. AdaLN is an adaptive layer norm, allowing the fusion to be conditioned on inputs $c$. A linear projection $W \in \mathbb{R}^{2d\times d}$ (and optionally, a bias) is necessary to ensure that $\tilde{x}$, composed of $a$ and $b$ concatenated in $x$ scaled and shifted by the adaptive layer norm, is first mixed before being split into input and gate to the GLU, respectively.

In the OAT model, the fusion layer is used on two occasions; when combining the NWP data with the topography, and when combining the uninformed tokens with the observations. $a$ and $b$ are those values respectively. In both cases, the conditioning value $c$ is an embedding of the day-of-year and hour-of-day informing the model about the time.

\subsection*{Geometry-aware attention:}

We consider three different attention mechanisms in this work. The first is very close to the standard dot product attention~\citep{vaswani2017attention}. It is given by Equation~\eqref{eq:attention}, but without attention bias and differs from the standard implementation, that only a set of neighbors $\mathcal{N}_i$ for each station $i$ is considered. The choice of neighborhood allows to introduce specific relational inductive biases in the architecture, similar to GNNs. We call this {\it no-geo} attention, since the attention mechanism is not provided with geographic information (apart from the topography that is embedded in the latent vectors $\mathbf z$). The second and main mechanism that is used throughout the paper if not specified otherwise we call {\it additive} and is given by Equation~\eqref{eq:attention}. The third attention mechanism we call {\it TE-TNP} since it is closest to the mechanism proposed by~\citet{ashman2024translation}, where the attention logits are given by
\begin{equation}
    \alpha_{i,j} = 
    \begin{cases}
    f\left(\left[Q_i^\top K_j \left\Vert \left(\boldx_i^{\rm elev}-\boldx_j^{\rm elev}\right.\right)\right]\right) & \text{ if } j\in \mathcal{N}_i, \\
    -\infty & \text{ otherwise },
    \end{cases}
\end{equation}
where the distance in position (and elevation) get concatenated to the dot product and then attention is formed non-linearly from these two sources. This is the richest attention class from a modeling perspective, but more difficult to interpret than the additive attention mechanism due to the non-linear mixing of $f$. The results of an ablation study for the different mechanisms can be found in Table~\ref{tab:att_oat_perf}.

Multi-head self-attention $\text{MHSA}_{\mathrm{geo}}(\boldz;\, \mathbf{x}_y^{\rm elev})$ projects the same embeddings $\boldz$ linearly into queries, keys and values. Multi-head cross-attention $\text{MHCA}_{\mathrm{geo}}(\boldz_{q}, \boldz_{k}, \boldz_{v};\, \mathbf{x}^{{\rm elev}, \star}, \mathbf{x}_y^{\rm elev})$, instead, projects distinct sequences into queries, keys and values. Throughout this work we assume that $\mathbf{z}_k=\mathbf{z}_v$ for cross-attention. The computation for self- and cross-attention is identical once the projections are formed, and we distinguish the two names purely for clarity in the following discussion.

\subsection*{Observation-assimilation Transformer Block} 

At observed locations we generate embeddings $\boldz_y$ informed by the observed $\boldy$, while at query location we have uninformed embeddings $\boldz^\star$. An encoder-decoder transformer performs a token mixing of those two sets:
\begin{equation}
    \boldz^\star_y={\rm Transformer}(\boldz^\star,\boldx^{{\rm elev},\star},\boldz_y,\boldx_y^{\rm elev}) = \text{Dec}((\boldz^\star,\boldx^{{\rm elev}, \star}), (\text{Enc}(\boldz_y,\boldx_y^{\rm elev}),\boldx_y^{\rm elev}))
\end{equation}
The architecture is close to the one described by~\citet{vaswani2017attention}, with residual connections, dropout, layer norm, and feed-forward block after the attention. The only differences are in that we do not have any positional embedding and we do not consider self-attention among the queries in the decoder blocks. Hence, the encoder is only composed of $\text{MHSA}_{\mathrm{geo}}$ blocks, while the decoder is only composed of $\text{MHCA}_{\mathrm{geo}}$ blocks.

The output of the final decoder layer is then used in the output head, where it is mapped to the parametric distribution via an MLP.
\begin{table}[t]
\centering
\small
\begin{tabular}{lll}
\toprule
\textbf{Component} & \textbf{Hyperparameter} & \textbf{Value} \\
\midrule
\multirow{4}{*}{NWP encoder}
  & Type                     & pointwise MLP, GELU \\
  & Input channels           & 4 (CH) / 3 (EU) \\
  & Hidden layers            & 1, width 16 \\
  & Output channels          & 16 \\
\midrule
\multirow{4}{*}{Topographic encoder}
  & Type                     & pointwise MLP, GELU \\
  & Input channels           & 5 (CH) / 2 (EU) \\
  & Hidden layers            & 1, width 16 \\
  & Output channels          & 16 \\
\midrule
Both encoders & Batch normalization & yes \\
\midrule
\multirow{5}{*}{Fusion layers}
  & Output / hidden channels & 32 / 32 \\
  & Conditioning dim.\ (time) & 4 \\
  & Normalization            & AdaLN (pre-norm) \\
  & Blocks                   & 1 \\
  & Dropout                  & 0.1 \\
\midrule
\multirow{9}{*}{OAT transformer}
  & Model dim.\ $\dmodel$    & 32 \\
  & Attention heads          & 4 \\
  & Encoder / decoder layers & 3 / 3 \\
  & Feedforward dim.         & 64 \\
  & Dropout                  & 0.1 \\
  & Geometric attention      & additive \\
  & Geo.\ bias input dim.    & 3 (CH) / 4 (EU) \\
  & Geo.\ bias FFN dim.      & 32 \\
  & Attention mask           & KNN, $k=8$ \\
\midrule
\multirow{3}{*}{Output head}
  & Input features           & 32 \\
  & Temperature              & Normal \\
  & Precipitation            & zero-inflated log-normal \\
\midrule
\multicolumn{2}{l}{Pre-OAT loss weight $w_\ell$} & 0.1 \\
\multicolumn{2}{l}{Trainable parameters} & $\approx 70$k \\
\bottomrule
\end{tabular}
\caption{{\bf OAT architecture hyperparameters.} CH: Switzerland dataset, EU: Europe dataset. Values are identical for both datasets except where noted.
The uninformed baseline has only the encoders, the NWP--topography fusion and the output head. It uses widths 32 in the encoders and 64 in the fusion layer and output head, and has $\approx 20$k parameters.}
\label{tab:oat-hyperparams}
\end{table}

\begin{table}[t]
\centering
\small
\begin{tabular}{lccc}
\toprule
\textbf{Setting} & \textbf{OAT / Uninf.} & \textbf{GP / DK-GP} & \textbf{Gridded TNP} \\
\midrule
Optimizer                   & AdamW & AdamW & Adam \\
Learning rate               & $10^{-3}$ & $10^{-2}$ (CH) / $5\times10^{-3}$ (EU) & $5\times10^{-4}$ \\
Weight decay                & $0.01$ & $0.01$ & --- \\
LR schedule                 & \multicolumn{2}{c}{cosine annealing, $T_{\max}=500$ epochs\textsuperscript{c}} & none \\
Batch size (time steps)\textsuperscript{a} & \multicolumn{2}{c}{$1024$ (CH) / $256$ (EU)\textsuperscript{c}} & $1$\textsuperscript{b} \\
Gradient clipping (norm)    & $1.0$ & $1.0$ & $0.5$ \\
Max.\ epochs                & $500$\textsuperscript{c} & $500$ & $300$ \\
Early-stopping patience     & $40$ & $40$ & --- \\
Float precision             & $32$ & $64$ & $32$ \\
Training loss               & CRPS & CRPS & CRPS \\
Model selection             & best val.\ loss & best val.\ CRPS (in data space) & best val.\ loss \\
Context/target split         & \multicolumn{3}{c}{$80\%$ / $20\%$ of training stations, resampled each step} \\
Hardware                    & \multicolumn{3}{c}{1 node, 4 GPUs (A100), DDP} \\
\bottomrule
\end{tabular}
\caption{{\bf Training configuration.} CH: Switzerland dataset, EU: Europe dataset.
\textsuperscript{a}Per GPU; effective batch size $\times4$. One sample is one hourly time step with all its stations.
\textsuperscript{b}The Gridded TNP generator emits one sample per step with $10\,000$ steps per epoch.
\textsuperscript{c}The region hold-out models use a batch size of $32$, $T_{\max}=100$ and at most $100$ epochs.}
\label{tab:training-hyperparams}
\end{table}
\begin{table}\centering
\begin{tabular}{l|ccc|ccc}
    \toprule
    & \multicolumn{3}{c|}{\textbf{Temperature}} & \multicolumn{3}{c}{\textbf{Precipitation}} \\
    \cmidrule(lr){2-4}\cmidrule(lr){5-7}
    \textbf{Method} & MAE $\downarrow$ & CRPS $\downarrow$ & Skill $\uparrow$ & MAE $\downarrow$ & CRPS $\downarrow$ & Skill $\uparrow$\\
    \midrule
    OAT (additive) & $1.046$ & $0.751$ & $0.045$ & $0.086$ & $0.064$ & $0.306$ \\
    OAT (vanilla) & $1.076$ & $0.771$ & $0.019$ & $0.088$ & $0.066$ & $0.287$ \\
    OAT (TE-TNP)  & $1.035$ & $0.743$ & $0.054$ & $0.085$ & $0.064$ & $0.310$ \\
    \bottomrule
\end{tabular}
\caption{{\bf Ablation of different attention mechanisms.} Evaluated on the Switzerland dataset.}
\label{tab:att_oat_perf}
\end{table}
\begin{sidewaystable}[p]
\centering
\resizebox{\textheight}{!}{%
\begin{tabular}{ll|ccc|ccc}
    \toprule
    & & \multicolumn{3}{c|}{\textbf{Temperature}} & \multicolumn{3}{c}{\textbf{Precipitation}} \\
    \cmidrule(lr){3-5}\cmidrule(lr){6-8}
    \textbf{Dataset} & \textbf{Method} & MAE $\downarrow$ & CRPS $\downarrow$ & Skill $\uparrow$ & MAE $\downarrow$ & CRPS $\downarrow$ & Skill $\uparrow$\\
    \midrule
    \multirow{6}{*}{\textbf{Switzerland}} & COSMO Interp. & [1.127, 1.464] & [1.132, 1.460] & [-0.785, -0.494] & [0.138, 0.155] & [0.138, 0.155] & [-0.633, -0.548] \\
    & Uninf. baseline & [0.968, 1.210] & [0.706, 0.879] & [0.000, 0.000] & [0.114, 0.131] & [0.086, 0.098] & [0.000, 0.000] \\
    & GP & [1.015, 1.203] & [0.751, 0.890] & [-0.108, 0.031] & [0.118, 0.143] & [0.101, 0.122] & [-0.278, -0.144] \\
    & DK-GP & [1.022, 1.230] & [0.760, 0.906] & [-0.136, 0.022] & [0.108, 0.134] & [0.092, 0.113] & [-0.158, -0.041] \\
    & Gridded TNP  & [1.236, 1.617] & [0.959, 1.293] & [-0.678, -0.186] & [0.090, 0.106] & [0.067, 0.079] & [0.180, 0.233] \\
    & OAT (ours) & [0.954, 1.143] & [0.684, 0.835] & [-0.013, 0.093] & [0.079, 0.094] & [0.058, 0.070] & [0.280, 0.331] \\
    \midrule
    \multirow{6}{*}{\textbf{Europe}} & ERA5 Interp. & [1.226, 1.379] & [1.233, 1.385] & [-0.492, -0.407] & [0.102, 0.109] & [0.102, 0.109] & [-0.640, -0.608] \\
    & Uninf. baseline & [1.190, 1.309] & [0.862, 0.945] & [-0.003, 0.003] & [0.079, 0.085] & [0.063, 0.067] & [-0.000, 0.000] \\
    & GP & [1.148, 1.234] & [0.906, 0.968] & [-0.059, -0.012] & [0.084, 0.090] & [0.073, 0.079] & [-0.180, -0.160] \\
    & DK-GP   & [1.101, 1.193] & [0.844, 0.906] & [0.004, 0.056] & [0.080, 0.086] & [0.071, 0.076] & [-0.153, -0.117] \\
    & Gridded TNP & [1.133, 1.223] & [0.842, 0.924] & [-0.012, 0.055] & [0.074, 0.079] & [0.058, 0.062] & [0.061, 0.083] \\
    & OAT (ours) & [0.954, 1.031] & [0.684, 0.750] & [0.185, 0.232] & [0.065, 0.069] & [0.050, 0.054] & [0.190, 0.216] \\
    \midrule
    \multirow{4}{*}{\shortstack[l]{\textbf{Europe} \\ (region hold-out)}} & ERA5 Interp. & [1.430, 2.035] & [1.423, 2.017] & [-0.492, -0.260] & [0.152, 0.168] & [0.152, 0.168] & [-0.719, -0.622] \\
    & COSMO Interp. & [1.049, 1.327] & [1.042, 1.315] & [-0.066, 0.171] & [0.132, 0.150] & [0.132, 0.151] & [-0.545, -0.400] \\
    & Uninf. baseline & [1.451, 1.997] & [1.054, 1.460] & [0.000, 0.000] & [0.110, 0.126] & [0.090, 0.102] & [0.000, 0.000] \\
    & OAT (ours) & [0.956, 1.260] & [0.686, 0.917] & [0.277, 0.449] & [0.084, 0.097] & [0.064, 0.075] & [0.240, 0.307] \\
    
    \bottomrule
\end{tabular}%
}
\caption{{\bf Bootstrapped confidence intervals for Table~\ref{tab:ml_results}}. 95 \% percentile intervals from a bootstrap over test stations (1000 resamples; each station's metric is averaged over the test period). MAE and CRPS intervals are marginal per model. Skill intervals are paired: model and uninformed baseline are evaluated on the same resampled stations in each resample. Stations are treated as independent, so the intervals do not account for spatial correlation.}
\label{tab:ci_table}
\end{sidewaystable}

\section{Gaussian Process baseline}
\label{app:gp}
For the GP models, we need a mean function and a covariance function, also called kernel. For the mean function, we use the same architecture as the uninformed baseline, except for the output head. As kernel, we use an automatic relevance determination kernel for the GP model, i.e. a radial basis function with dimension specific length scales. The input to the kernel are location and elevation. For the DK-GP~\citep{wilson2016deep,zanetta2026efficient}, we use the same kernel multiplied by an RBF kernel on a 4-dimensional MLP embedding of the feature trunk output, which combines NWP, topography, and time (2 hidden layers and a linear readout, width 64, ReLU). 
The loss is the CRPS on training held out stations. Context stations and target stations, on which the loss is computed, are sampled in the same manner as in the OAT training. For precipitation, we model the variable in log-space with a small offset to account for the dry events. Compared to OAT training, we have to resort to 64-bit precision to ensure numerical stability.
The hyperparameters of both GP variants are given in Table~\ref{tab:gp-hyperparams}.

\begin{table}[t]
\centering
\small
\begin{tabular}{lll}
\toprule
\textbf{Component} & \textbf{Hyperparameter} & \textbf{Value} \\
\midrule
\multirow{2}{*}{Feature trunk}
  & Architecture             & as uninformed baseline, no output head \\
  & Embedding dim.           & 64 \\
\midrule
Mean function & MLP on embedding & $64 \to 64 \to 1$, ReLU \\
\midrule
\multirow{4}{*}{Stationary kernel}
  & Type                     & RBF with ARD \\
  & Inputs (CH)              & easting, northing, elevation \\
  & Inputs (EU)              & longitude, latitude, elevation \\
  & Lower bounds             & length scale $10^{-2}$, output scale $10^{-2}$ \\
\midrule
\multirow{3}{*}{Deep kernel (DK-GP only)}
  & Type                     & RBF on MLP features, fixed scales \\
  & MLP on embedding         & $64 \to 64 \to 64 \to 4$, ReLU \\
  & Combination              & product with stationary kernel \\
\midrule
\multirow{2}{*}{Likelihood}
  & Type                     & Gaussian, one model per variable \\
  & Noise variance           & init.\ $1.0$, lower bound $10^{-3}$ \\
\midrule
\multirow{2}{*}{Targets}
  & Transform                & standardized \\
  & Precipitation            & $\log(y + 10^{-3})$ before standardization \\
\midrule
\multicolumn{2}{l}{Trainable parameters (CH)} & GP $\approx 28$k, DK-GP $\approx 45$k \\
\bottomrule
\end{tabular}
\caption{{\bf GP and DK-GP hyperparameters.} Values are identical for both GP variants and for the Switzerland (CH) and Europe (EU) datasets except where noted. Kernel inputs are standardized. For the training configuration see Table~\ref{tab:training-hyperparams}.}
\label{tab:gp-hyperparams}
\end{table}

\section{Gridded TNP baseline}
\label{app:gridded-tnp}
We use the gridded TNP of~\citet{ashman2025gridded} based on the implementation of the authors. The NWP field is subsampled with a fixed stride and encoded together with the observations onto a latent pseudo-token grid. Swin attention mixes the latent grid, and a cross-attention decoder reads it out at the query locations. As described in the main text, we give the decoder the same topographic features as OAT. The hyperparameters are given in Table~\ref{tab:gridded-tnp-hyperparams}.

\begin{table}[t]
\centering
\small
\begin{tabular}{llc}
\toprule
\textbf{Component} & \textbf{Hyperparameter} & \textbf{Value} \\
\midrule
\multirow{3}{*}{Latent grid}
  & NWP stride               & 10 (CH) / 4 (EU) \\
  & Pseudo-token grid        & $80\times120$ (CH) / $40\times72$ (EU) \\
  & NWP input channels       & 4 (CH) / 3 (EU) \\
\midrule
\multirow{2}{*}{Position encoding}
  & Type                     & spherical harmonics (absolute lon./lat.) \\
  & Legendre polynomials / embedding dim. & 10 / 128 \\
\midrule
\multirow{2}{*}{Token encoders}
  & Observation / NWP MLP    & 2 layers, width 128 \\
  & Topographic MLP          & 5 (CH) / 2 (EU) $\to$ 32, batch norm \\
\midrule
\multirow{5}{*}{Transformer}
  & Embedding dim.           & 128 \\
  & Heads $\times$ head dim. & $8\times16$ \\
  & Layers                   & 5 \\
  & Feedforward dim.         & 128 \\
  & Swin window / shift      & $4\times4$ / $2\times2$ \\
\midrule
Grid decoder & Nearest pseudo-tokens per query ($k$) & 9 \\
\midrule
\multirow{2}{*}{Output head}
  & Decoder MLP              & 2 layers, width 128 \\
  & Distributions            & as OAT (Normal / zero-inflated log-normal) \\
\midrule
\multicolumn{2}{l}{Trainable parameters} & $\approx 2.5$M (CH) / $\approx 1.6$M (EU) \\
\bottomrule
\end{tabular}
\caption{{\bf Gridded TNP hyperparameters.} CH: Switzerland dataset, EU: Europe dataset. For the training configuration see Table~\ref{tab:training-hyperparams}.}
\label{tab:gridded-tnp-hyperparams}
\end{table}

\section{Evaluation of model calibration}
\label{app:calibration}
We consider $500$ predictive samples on all test stations over the whole test year. For each station and time step, the PIT value is the fraction of predictive samples that fall at or below the observation, and the histogram of these values across all pairs is uniform when the forecasts are calibrated. For the reliability diagram, we take the predicted probability of a dry event as the fraction of zero-valued samples as well as the actual binary outcome, bin the pairs by the predicted probability, and plot the mean predicted probability against the observed frequency of dry observations in each bin, which lies on the diagonal under perfect calibration.

\section{Model Scalability} \label{app:scalability}

\begin{figure}
     \centering
     \begin{subfigure}[b]{0.45\textwidth}
         \centering
         \includegraphics[width=\textwidth]{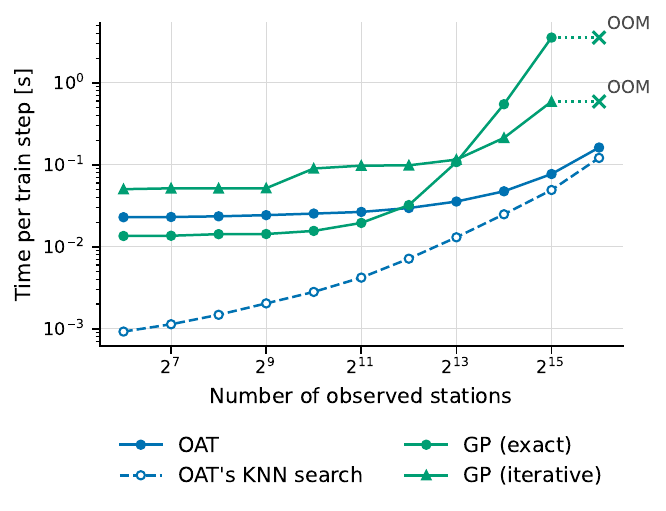}
         \caption{Training with fixed number of query points.}
         \label{fig:scaling_fixed_training}
     \end{subfigure}
     \hfill
     \begin{subfigure}[b]{0.45\textwidth}
         \centering
         \includegraphics[width=\textwidth]{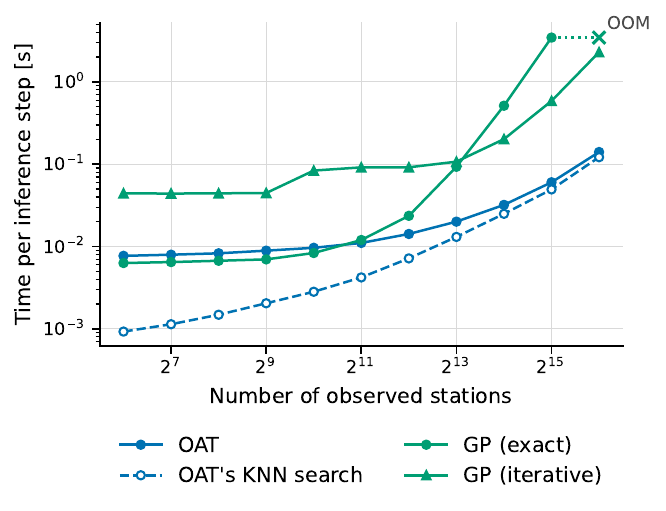}
         \caption{Inference with fixed number of query points.}
         \label{fig:scaling_fixed_inference}
     \end{subfigure}
     \begin{subfigure}[b]{0.45\textwidth}
         \centering
         \includegraphics[width=\textwidth]{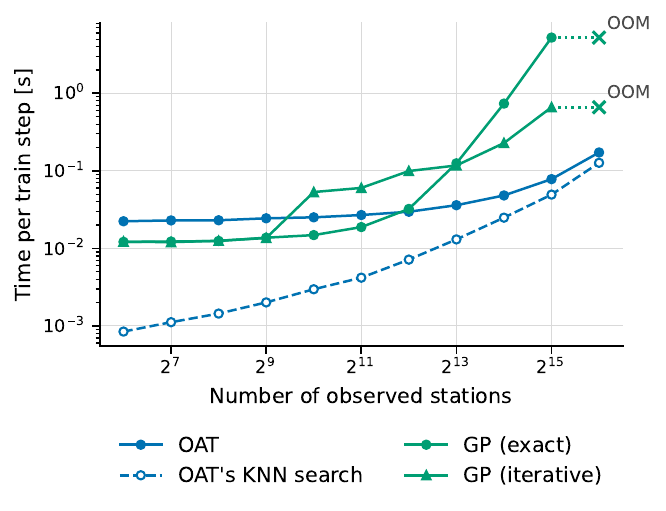}
         \caption{Training with proportional number of query points.}
         \label{fig:scaling_proportional_training}
     \end{subfigure}
     \hfill
     \begin{subfigure}[b]{0.45\textwidth}
         \centering
         \includegraphics[width=\textwidth]{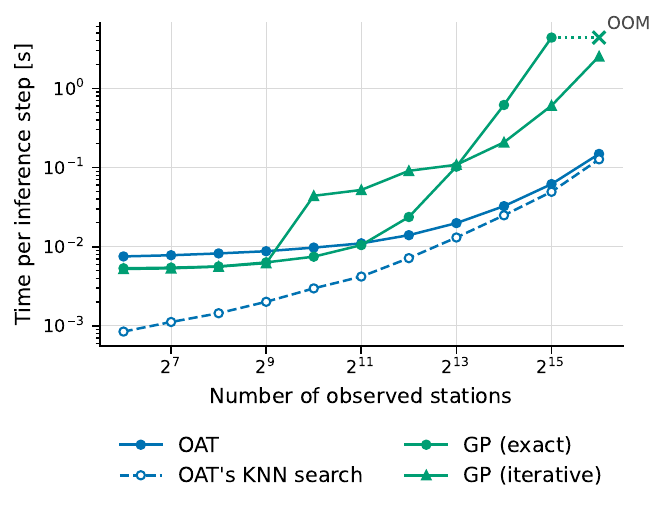}
         \caption{Inference with proportional number of query points.}
         \label{fig:scaling_proportional_inference}
     \end{subfigure}
        \caption{Scalability analysis of OAT and GP. We empirically measure the time per step under two scenarios (the rows in the figure) as the number of observed stations grows up to $2^{16} \approx 65'000$. In the first scenario, there is a fixed number of 1024 query points, while the second scenario has a number of query points proportional to the observed stations to maintain a 80\%-20\% ratio.
        The columns show the results for a training step, which includes forward and backward pass, and an evaluation step.
        A major bottleneck of the OAT's forward pass is the nearest neighbour computation, which we include in the plot as a dashed line. For the GP, we compare two ways of computing its predictive distribution: GP (exact) relies on a Cholesky decomposition, whose cost grows cubically with the number of points, while GP (iterative) uses GPyTorch's default, which switches from Cholesky to iterative conjugate-gradient and Lanczos methods above 800 points.}
        \label{fig:scaling}
\end{figure}

\begin{figure}
     \centering
     \begin{subfigure}[b]{0.45\textwidth}
         \centering
         \includegraphics[width=\textwidth]{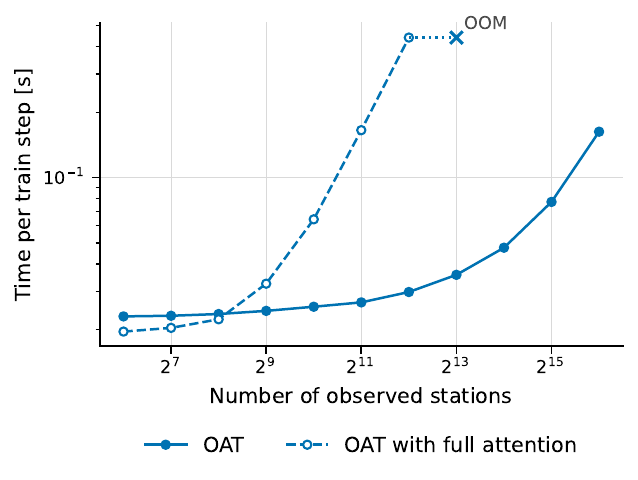}
         \caption{Training with fixed number of query points.}
         \label{fig:fullattention_fixed_train}
     \end{subfigure}
     \hfill
     \begin{subfigure}[b]{0.45\textwidth}
         \centering
         \includegraphics[width=\textwidth]{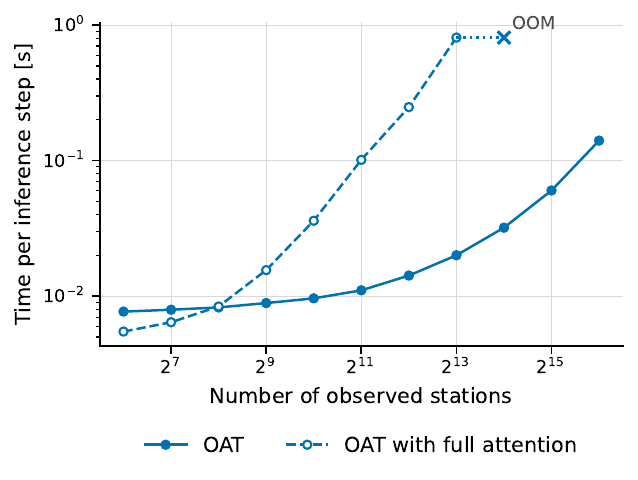}
         \caption{Inference with fixed number of query points.}
         \label{fig:fullattention_fixed_inference}
     \end{subfigure}
     \begin{subfigure}[b]{0.45\textwidth}
         \centering
         \includegraphics[width=\textwidth]{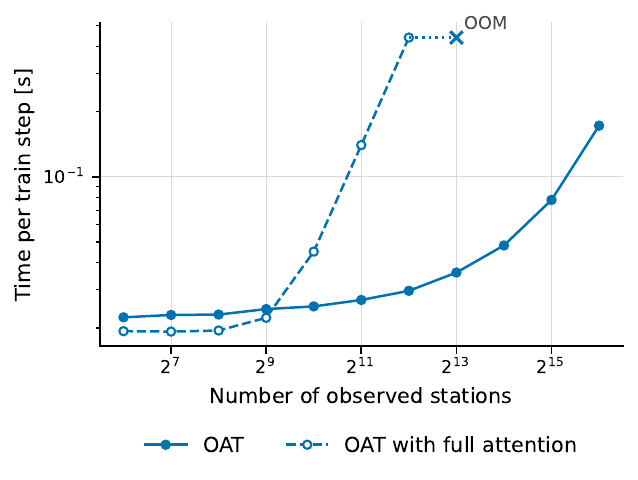}
         \caption{Training with proportional number of query points.}
         \label{fig:fullattention_prop_train}
     \end{subfigure}
     \hfill
     \begin{subfigure}[b]{0.45\textwidth}
         \centering
         \includegraphics[width=\textwidth]{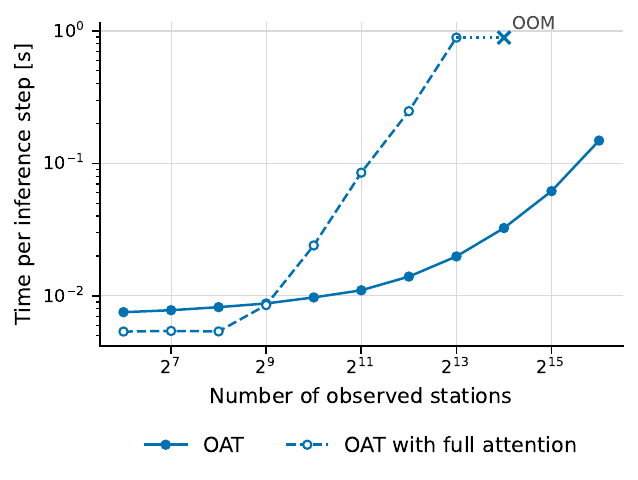}
         \caption{Inference with proportional number of query points.}
         \label{fig:fullattention_prop_inference}
     \end{subfigure}
        \caption{Comparison between a full-attention scheme and the default sparse attention scheme used in the OAT model to perform training and inference steps. The OAT setting is the same as in Figure \ref{fig:scaling}, with the only difference that the full-attention scheme is turned on.}
        \label{fig:fullattention_vs_sparse}
\end{figure}

We showcase empirical scalability comparing the GP and OAT models in Figure \ref{fig:scaling}. The model's hyperparameters are taken from the Switzerland experiment with a small change for the OAT. Because the GP is univariate, for the sake of this experiment, we make OAT univariate as well. We evaluate the models with a fixed number of 1024 query stations as well as maintaining a proportional number of query stations as the number of conditioning stations grows. In the latter setting, the 80-20 ratio is used. The deep GP model's scalability was comparable to the standard GP, and is therefore omitted from the plots.
All timings were measured on a single NVIDIA A100 GPU. We discard 3 warm-up forward passes and report the mean time over the following 10.
In the proportional setting, the exact and approximate GP have the same speed before $2^{10}=1024$ observed stations. This is because, by default, the approximate solver is only called when there are more than $800$ conditioning points. For the exact GP, we manually increase this limit to be greater than any number of points in the experiment. In the setting with fixed 1024 observed stations, the approximate solver is always in play. Interestingly, the exact solver is faster for such small matrices. We notice that for large values of number of points, the computation time of the OAT is dominated by the KNN computation. Under the assumption that all observed points are known a priori, a single pre-computed KD-tree could be used. Approximate KNN algorithms could also be employed to speed up OAT's training and inference~\citep{ma2021sub}. 
The main takeaway from this scalability analysis is that, in this setting, the GP model is faster for small numbers of points but scales worse as the number of points grows. We attribute its early advantage to the highly optimized low-level linear algebra routines in GPyTorch~\citep{gardner2018gpytorch} and see a clear advantage of OAT for larger datasets.
Figure \ref{fig:fullattention_vs_sparse} shows that while the KNN computation is a dominant computational component, it is still beneficial over a full attention scheme.


\end{document}